\documentclass{article} 
\usepackage{iclr2027_conference}
\iclrfinalcopy

\usepackage{amsmath,amsfonts,bm}

\def\eqref#1{equation~\ref{#1}}

\def\1{\bm{1}}

\DeclareMathAlphabet{\mathsfit}{\encodingdefault}{\sfdefault}{m}{sl}
\SetMathAlphabet{\mathsfit}{bold}{\encodingdefault}{\sfdefault}{bx}{n}

\usepackage{times}
\usepackage{hyperref}
\usepackage{url}

\usepackage{booktabs}
\usepackage{array}
\usepackage{pifont}
\usepackage{graphicx}
\usepackage{subcaption}
\usepackage{placeins}
\usepackage{multirow}
\usepackage{colortbl}
\usepackage{listings}
\usepackage{tabularx}
\usepackage{enumitem}
\usepackage{minitoc}

\newcommand{\cmark}{\ding{51}}

\newcolumntype{L}[1]{>{\raggedright\arraybackslash}p{#1}}
\newcolumntype{K}[1]{>{\centering\arraybackslash}p{#1}}

\title{From Perception to Prediction and Intervention: Benchmarking Proactive Video \\
Assistants in Real-Life Scenarios}

\author{
\parbox[t]{\dimexpr\textwidth-2\tabcolsep\relax}{
\textbf{Boyu Cui$^{1,*}$, Junhao Luo$^{1,*}$, Zhihao Wang$^{1,*}$, Linghan Chen$^{1,*}$, Zejun Li$^{1,*}$, Jiwen Zhang$^{1}$}\\[3pt]
\textbf{Siyuan Wang$^{3,\dagger}$, Zhongyu Wei$^{1,2,\dagger}$}\\[6pt]
\normalfont
$^{1}$Fudan University\quad $^{2}$Shanghai Innovation Institute\quad $^{3}$The Chinese University of Hong Kong\\[3pt]
{\small $^{*}$Equal contribution\qquad $^\dagger$Corresponding author}
}
}

\begin{document}

\maketitle
\doparttoc
\faketableofcontents
\lhead{Technical Report}  

\begin{abstract}
Recent multimodal large language models have advanced video understanding, but existing benchmarks mainly evaluate whether models can interpret observed visual content. In real-world scenarios, assistants must additionally anticipate what assistance is needed and decide when to provide it. To address this gap, we introduce \textsc{P$^3$IBench}, a benchmark for proactive video understanding in real-life streaming scenarios. It evaluates three complementary capabilities: daily perception, risk prediction and intervention. We benchmark 12 representative open-source and proprietary video-language models, including streaming and adapted non-streaming systems. Our evaluation shows that current models struggle to make well-calibrated response decisions. Finally, we construct additional proactive training data and train \textsc{P$^3$I-Assistant-4B}, achieving improvements on several tasks, though gains remain uneven across capabilities. Together, these findings show that effective proactive assistance requires not only visual understanding, but also precise response timing and appropriate restraint.
\end{abstract}

\begin{center}
\small
\href{https://kusu97.github.io/P3IBench}{\textbf{Homepage}}
\quad$\vert$\quad
\href{https://github.com/Cuibing-fundemental/P3IBench}{\textbf{Code}}
\quad$\vert$\quad
\href{https://huggingface.co/datasets/AstroCuibing/P3IBench}{\textbf{Dataset}}
\end{center}

\section{Introduction}

Streaming video enables intelligent assistants to continuously observe evolving
real-world environments and interact with users beyond offline or query-triggered
video understanding. A key capability is proactive streaming
video understanding, where an assistant autonomously determines whether and
when to respond without explicit user queries~\citep{wang2025proactivevideoqa,zhao2026omnipro,du2026liveprobench}.
Such assistants must not only
understand ongoing events, but also respond at the right moment for their assistance
to remain useful. This creates a growing need to reliably evaluate both the
appropriateness and timeliness of proactive responses~\citep{wang2025proactivevideoqa,du2026liveprobench}.


Existing benchmarks have advanced the evaluation of proactive video
understanding~\citep{lin2024streamingbench,li2025ovobench,wang2025proactivevideoqa,shi2026river}, but two key limitations remain. First, they primarily focus on \emph{perception},
such as recognizing ongoing events or predefined conditions. Real-life proactive
assistance should go beyond perception to \emph{predict} emerging events or risks
early enough for advance warning, and \emph{intervene} with timely reminders,
corrections, or guidance toward a user's goal~\citep{zeng2025streamforest,ran2026egopro,zhao2026omnipro}.
These capabilities require models
to reason not only about what is currently observable, but also about what may
happen next and when guidance becomes necessary. Second, expanding to prediction and intervention poses stricter requirements on benchmark data quality, especially the precision of temporal annotations~\citep{wang2025proactivevideoqa,ran2026egopro}.
A proactive response is useless if delayed, yet ungrounded if triggered before sufficient evidence accumulates. Evaluating such systems thus demands precise, fine-grained temporal boundaries for when a response becomes both justified and actionable.

\begin{figure}[t]
\setlength{\abovecaptionskip}{0pt}
    \centering
    \begin{subfigure}[t]{0.48\textwidth}
        \centering
        \includegraphics[width=\linewidth]{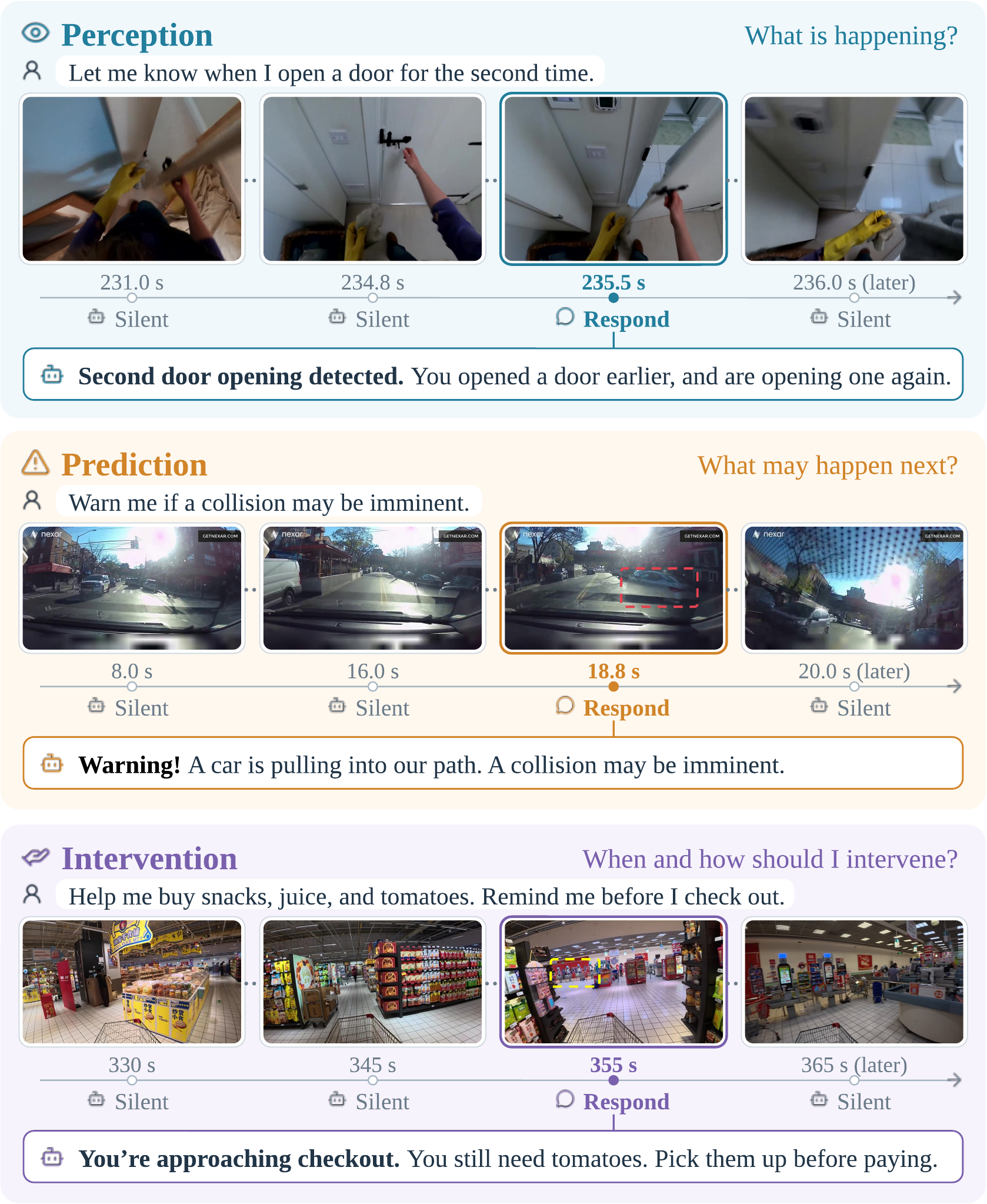}
        \caption{Three proactive capabilities.}
        \label{fig:teaser-capabilities}
    \end{subfigure}
    \hfill
    \begin{subfigure}[t]{0.51\textwidth}
        \centering
        \includegraphics[width=\linewidth]{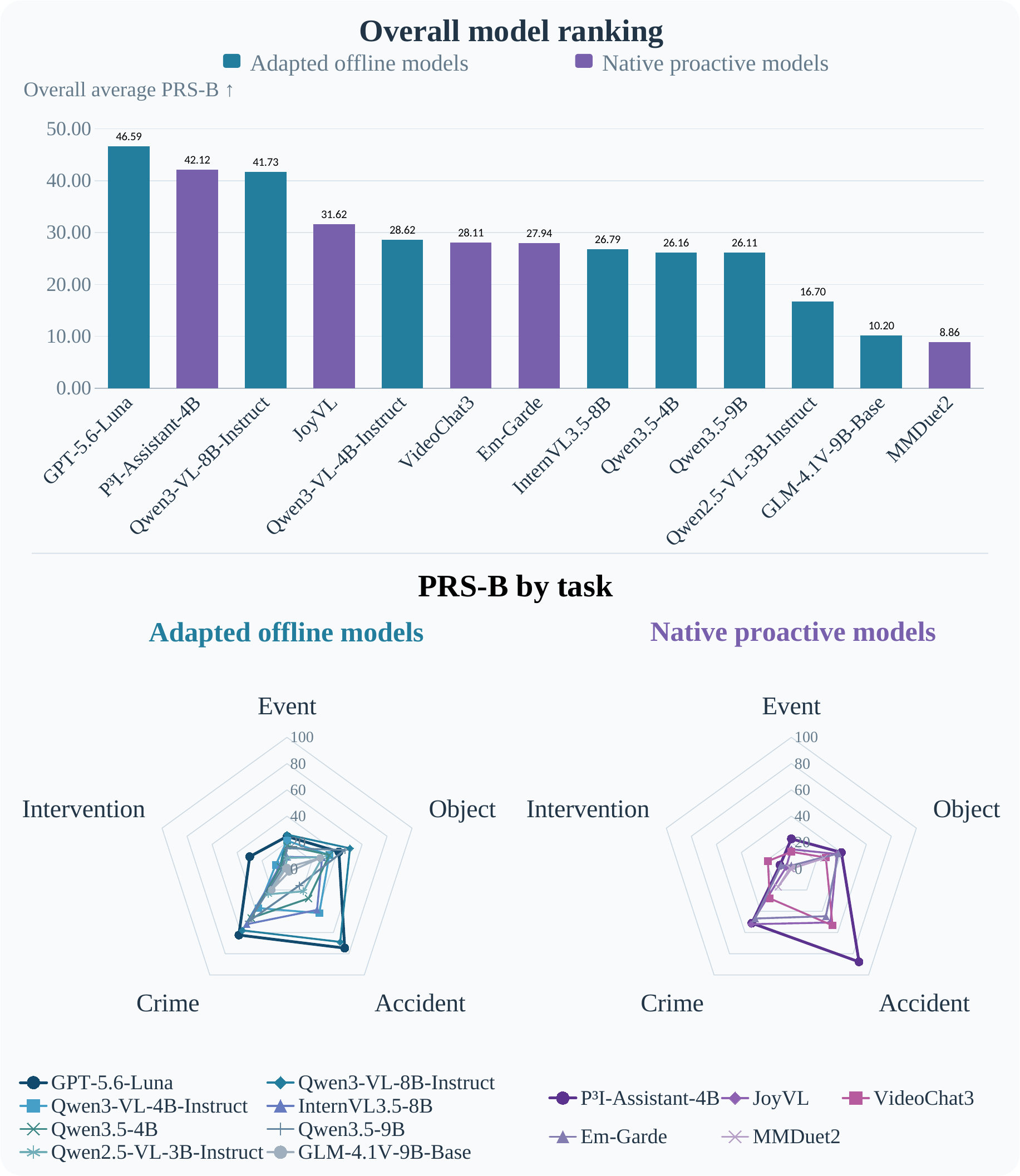}
        \caption{Overall model ranking.}
        \label{fig:teaser-ranking}
    \end{subfigure}

    \caption{
        \textbf{Overview of \textsc{P$^3$IBench} and model performance.}
        (a) Three complementary proactive capabilities:
        \emph{Perception}, recognizing relevant events as they occur;
        \emph{Prediction}, anticipating risks before they materialize;
        and \emph{Intervention}, proactively providing assistance or
        corrections when needed.
        (b) Overall ranking of evaluated models by average PRS-B (higher is better),
        together with task-wise PRS-B profiles across five subtasks.
        Detailed numerical results are reported in Table~\ref{tab:main_results}.
    }
    \label{fig:teaser}
\end{figure}


To fill these gaps, we introduce \textsc{P$^3$IBench}
(\textbf{Proactive Perception-Prediction-Intervention}), a benchmark for
evaluating proactive video assistants in diverse real-life scenarios.
P$^3$IBench contains 533 video-query samples from 429 unique videos, organized
into three subsets: \emph{Daily Perception} for monitoring events and objects in
everyday scenarios, \emph{Risk Alert} for anticipating emerging hazards in traffic and crime-related scenarios, and \emph{Event Intervention} for providing timely, context-dependent guidance toward user goals.
Each sample pairs an initial instruction with temporally
grounded response targets, enabling evaluation of both response correctness and
timeliness. To construct reliable evaluation data, we develop a hybrid
data construction pipeline that combines multi-LLM annotation with human
annotation and review, with particular emphasis on precisely identifying when
each response becomes justified and useful.

Using \textsc{P$^3$IBench}, we evaluate 12 representative models spanning open-source and proprietary systems, including four native proactive streaming models and eight general-purpose offline MLLMs adapted to causal streaming inference. We introduce two metrics, PRS-U and PRS-B, that reward correct responses at appropriate times, while accounting for missed responses when
models fail to respond and penalizing unnecessary or repeated outputs, and analyze performance across capabilities, temporal factors, and streaming inference settings.
Overall, no model performs reliably across all three proactive capabilities, with the strongest system reaching only 46.59 PRS-B (Figure~\ref{fig:teaser-ranking}). 
Models within each category exhibit consistent but distinct failure patterns:
adapted offline models tend to over-respond with unnecessary or repeated outputs,
whereas native proactive models are often overly conservative, missing warranted responses or responding with substantial delay. These results show that existing video understanding and streaming models remain insufficient for proactive assistance, which additionally requires well-calibrated decisions about whether and when to respond.
To this end, we further construct proactive training data and train
\textsc{P$^3$I-Assistant-4B} on top of Qwen3-VL-4B via supervised fine-tuning and reinforcement
learning, achieving clear gains on \textsc{P$^3$IBench}. In summary, our contributions are threefold:
\begin{itemize}[leftmargin=1em]
    \item We introduce \textsc{P$^3$IBench}, a benchmark for proactive video
    assistance that jointly evaluates \emph{perception}, \emph{prediction}, and
    \emph{intervention} in real-life streaming scenarios, with high-quality
    response targets and temporal annotations through a
    hybrid multi-LLM and human annotation pipeline.

    \item We systematically evaluate 12 representative models and reveal clear
    limitations of current approaches, including distinct failure patterns
    across model categories in over-response, missed responses, and response
    latency, along with their sensitivity to temporal and streaming settings.

    \item We construct proactive training data and train
    \textsc{P$^3$I-Assistant-4B} through supervised fine-tuning and reinforcement
    learning, achieving clear improvements on \textsc{P$^3$IBench} and
    demonstrating the effectiveness of targeted training for proactive video
    assistance.
\end{itemize}

\begin{table}[t]
\centering
\caption{Comparison of representative benchmarks for streaming video understanding and proactive assistance. Pe, Pr, and I indicate perception, prediction of not-yet-observed future events or risks, and proactive intervention, respectively. HR denotes human review.}
\label{tab:benchmark_comparison}
\vspace{-4pt}
\small
\setlength{\tabcolsep}{3.0pt}












\resizebox{0.9\textwidth}{!}{
\begin{tabular}{lcccccccccc}
\toprule
\textbf{Benchmark}
& \multicolumn{2}{c}{\textbf{Scale}}
& \multicolumn{3}{c}{\textbf{Scenarios}}
& \multicolumn{3}{c}{\textbf{Capabilities}}
& \multicolumn{2}{c}{\textbf{Quality Control}} \\
\cmidrule(lr){2-3}
\cmidrule(lr){4-6}
\cmidrule(lr){7-9}
\cmidrule(lr){10-11}

&
\textbf{\#Videos}
& \textbf{\#Samples}
& \textbf{Daily}
& \textbf{Traffic}
& \textbf{Crime}
& \textbf{Pe}
& \textbf{Pr}
& \textbf{I}
& \textbf{Anno}
& \textbf{HR} \\
\midrule

StreamingBench
& 900
& 4,500
& \cmark &  & 
& \cmark & \cmark & 
& LLM+Human
& \cmark \\

OVO-Bench
& 644
& 2,814
& \cmark &  & 
& \cmark & \cmark & 
& LLM+Human
& \cmark \\

ProactiveVideoQA
& 1,377
& 1,427
& \cmark &  & \cmark
& \cmark &  & 
& LLM+Rule+Human
&  \\

ODV-Bench
& 1,190
& 6,348
&  & \cmark & 
& \cmark & \cmark & 
& LLM+Rule+Human
& \cmark \\

RIVER
& 1,067
& 4,278
& \cmark &  & 
& \cmark &  & 
& LLM+Rule
& \cmark \\

EgoPro-Bench
& 2,400
& 2,400
& \cmark &  & 
& \cmark &  & \cmark
& LLM
& \cmark \\

OmniPro
& 1,262
& 2,700
& \cmark &  & 
& \cmark &  & \cmark
& LLM
& \cmark \\

LiveProBench
& 911
& 911
& \cmark &  & 
& \cmark &  & 
& Human
& \cmark \\

\midrule

\textbf{P$^3$IBench (ours)}
& 429
& 533
& \cmark & \cmark & \cmark
& \cmark & \cmark & \cmark
& LLM+Human
& \cmark \\

\bottomrule
\end{tabular}
}

\end{table}

\section{Related Works}
\label{sec:related-work}

\subsection{Online and Proactive Video Models}
Online video MLLMs extend offline video understanding to causally arriving streams, while proactive variants additionally determine when accumulated evidence warrants a response. VideoLLM-online~\citep{chen2024videollmonline} introduces temporally aligned streaming dialogue through a causal interleaved video--language interface, and MMDuet~\citep{wang2025mmduet} later adopts a video--text duet format for time-sensitive interaction. MMDuet2~\citep{wang2026mmduet2} further advances this line by learning response-or-silence decisions with multi-turn reinforcement learning. Other representative designs include VideoChat-Online~\citep{huang2025videochatonline}, which introduces a Pyramid Memory Bank for maintaining online context; Dispider~\citep{qian2025dispider}, which separates perception, decision, and reaction; and Em-Garde~\citep{zheng2026emgarde}, which uses lightweight proposal matching for efficient response triggering. Recent systems such as VideoChat3~\citep{li2026videochat3} and JoyAI-VL-Interaction~\citep{yao2026joyai} further support proactive response control in continuous video streams.

\subsection{Streaming and Proactive Video Benchmarks}
Existing benchmarks differ substantially in their coverage of perception, prediction, and intervention, as summarized in Table~\ref{tab:benchmark_comparison}. StreamingBench~\citep{lin2024streamingbench} and OVO-Bench~\citep{li2025ovobench} evaluate online perception together with prediction-oriented tasks, while ODV-Bench~\citep{zeng2025streamforest} similarly covers perception and prediction with a focus on driving scenarios. ProactiveVideoQA~\citep{wang2025proactivevideoqa}, RIVER~\citep{shi2026river}, and LiveProBench~\citep{du2026liveprobench} primarily emphasize perception-centered interaction. In contrast, EgoPro-Bench~\citep{ran2026egopro} extends evaluation toward personalized proactive assistance in egocentric activities, while OmniPro~\citep{zhao2026omnipro} includes procedural guidance through its Sequential Step Instruction sub-task. Overall, prior benchmarks typically cover only a subset of \emph{Perception}, \emph{Prediction}, and \emph{Intervention}, or study their combination within specialized settings.

Beyond capability coverage, proactive evaluation places particularly strong demands on annotation quality and temporal precision. As summarized in Table~\ref{tab:benchmark_comparison}, prior benchmarks adopt diverse construction pipelines, including LLM- or rule-assisted annotation with varying degrees of human review, as well as fully human annotation. While these approaches support scalable benchmark construction, they often inherit heterogeneous source annotations or rely on automatically generated intermediate labels, making consistent control over response validity and temporal boundaries challenging.

\textsc{P$^3$IBench} addresses these limitations through task-specific construction and verification procedures. Compared with prior benchmarks that often rely on heterogeneous source annotations or limited post-hoc review, our pipeline combines automated annotation with targeted verification and human inspection to provide more consistent, high-confidence response targets. At the same time, \textsc{P$^3$IBench} jointly evaluates \emph{Perception}, \emph{Prediction}, and \emph{Intervention} under a unified causal protocol across everyday monitoring, anticipatory risk warning, and goal-conditioned task assistance.

\section{\texorpdfstring{P$^3$IBench}{P3IBench}}
\subsection{Benchmark Construction}
\label{sec:task_taxonomy}

\textsc{P$^3$IBench} evaluates proactive video understanding: given a task
instruction and an incrementally observed video stream, a model must
decide when and what to respond using only observations available up to
the current time, while remaining silent when no task-relevant response
is warranted. We organize the benchmark into three categories:
\textbf{Daily Perception}, \textbf{Risk Alert}, and
\textbf{Event Intervention}, corresponding broadly to perception,
prediction, and intervention.

\textbf{Daily Perception.}
This category evaluates instruction-conditioned monitoring of events and object states. It includes \textit{Event Monitoring} task and \textit{Object Monitoring} task. 
Event Monitoring considers events at three levels of granularity: a single specified atomic event (e.g., \textit{wash plates}), a compositional process formed by multiple temporally related events (e.g., \textit{wash plates, then dry them}), and a semantic event category that may correspond to multiple concrete atomic events (e.g., \textit{wash something}). 
Object Monitoring requires tracking a specified object and reporting when it disappears from view. For tasks in this category, the response should be issued as soon as the instructed event or object state becomes reliably identifiable.

To enhance the diversity of real-world daily activities, we construct 
Event Monitoring from Ego4D~\citep{Grauman_2022_CVPR}, using its dense timestamped narrations to generate candidate queries, match them to relevant events with visual evidence, and localize their temporal boundaries. Two independent vision-language models are used for boundary estimation, followed by agreement checking, verification, and confidence-based filtering (Details can be found in Appendix~\ref{app:task_definitions_examples}).
Object Monitoring is constructed from TAG~\citep{zhang2026sam2trackinggranularity}, where segmentation masks are used to identify object disappearance events and their timestamps. 
The resulting instances are additionally inspected or manually refined to reduce event-identity and temporal-boundary errors.

\textbf{Risk Alert.}
This category evaluates timely warnings about emerging or ongoing hazards, with particular emphasis on warning timing, which is a capability rarely considered in prior video-understanding benchmarks. 
Specifically, it includes \textit{Accident Warning} task and \textit{Crime Alert} task. 
Accident Warning requires the model to recognize an imminent collision risk and issue a warning before the collision occurs. We derive it with driving videos from Nexar~\citep{nexar2025dashcamcollisionprediction}
dataset using its original accident annotations. 
Crime Alert requires a warning once criminal or suspicious behavior becomes detectable, and is constructed from UCF-Crime~\citep{yuan2023surveillancevideoandlanguageunderstandingnew} dataset.
The selected accident annotations are manually checked and the coarse
crime intervals are manually refined to obtain more accurate temporal
boundaries. 

\textbf{Event Intervention.}
This category evaluates proactive assistance conditioned on the user's
goal and observed task progress. The model must identify situations
that warrant assistance and provide a relevant reminder, correction, or
guidance while remaining useful. Since multiple situations may arise
within one video, the model may need to intervene several times without
additional user queries.
We construct this category from first-person videos recorded in five
real-world scenarios: grocery shopping, navigation, procedural
maintenance, desk organization, and item packing. Each video follows a
manually designed task script that contains at least one intentional error
in step order, task content, or both. We then annotate the temporal
windows in which an intervention is visually grounded and useful and
pair each window with a reference response describing the essential
object, location, operation error, or next action. All scripts, videos,
instructions, intervention windows, and reference responses are
manually annotated and verified.

Together, the three categories capture the
\textit{Perception--Prediction-Intervention} theme of P$^3$IBench:
recognizing relevant events and object states, warning about hazards at
the appropriate time, and providing goal-directed assistance from
streaming observations.

\subsection{Benchmark Statistics}
\begin{figure}[t]
\setlength{\abovecaptionskip}{0pt}
    \centering
    \includegraphics[width=\textwidth]{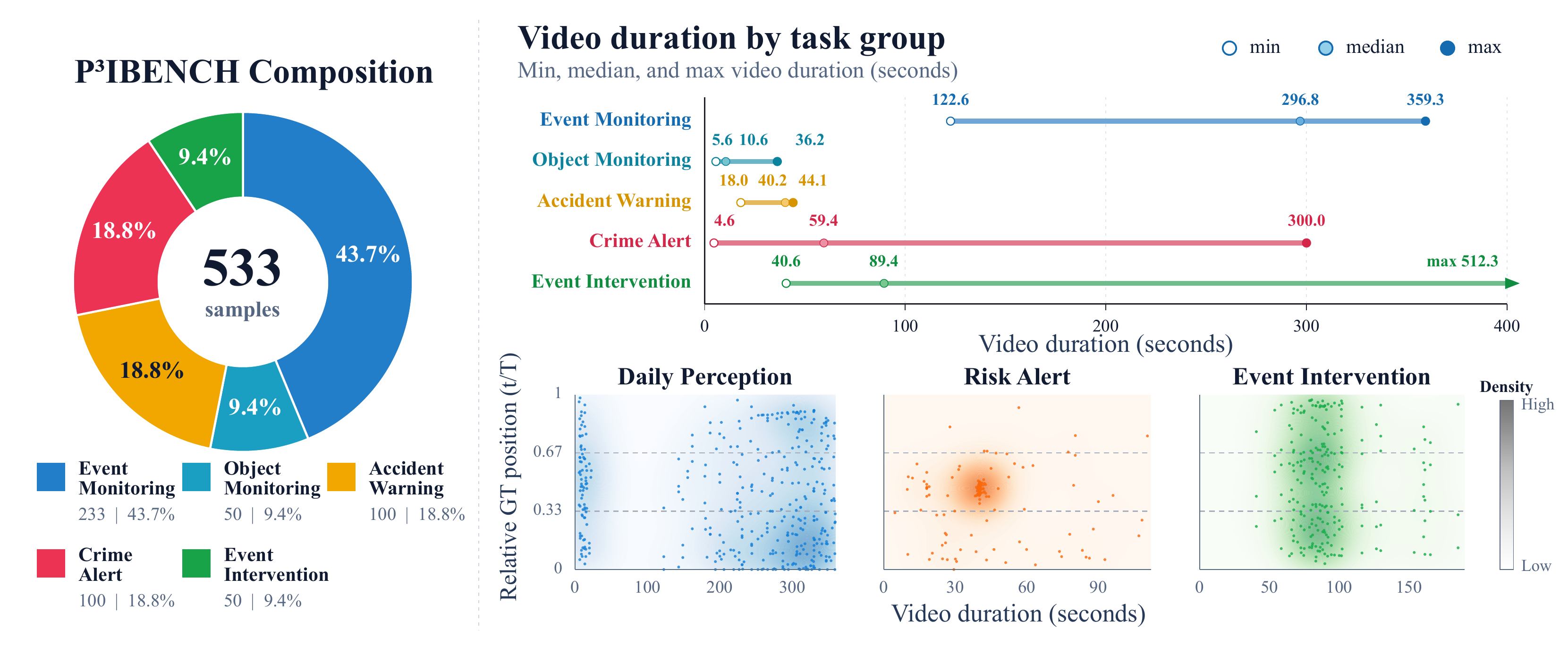}
    \centering
    \caption{\textbf{Benchmark Statistics.} 
    (Left) Task distribution. (Top right) Video duration distribution across tasks. 
    (Bottom right) Distribution of GT trigger positions across categories, normalized by video duration; dashed lines divide the video timeline into early, middle, and late thirds. For readability, the Risk Alert and Event Intervention scatterplots show GT events from videos up to 112s and 190s, respectively, covering 90.2\% and 90.7\% of their GT events. 
    }
    \label{fig:composition-duration}
\end{figure}

\textsc{P$^3$IBench} contains 533 query samples from 429 unique videos across three categories and five task groups, as detailed in the left side of Figure~\ref{fig:composition-duration}. The top right panel of Figure~\ref{fig:composition-duration} demonstrates that \textsc{P$^3$IBench} covers substantially different temporal scales across tasks. Median video durations range from 10.6 to 296.8 seconds, while the longest Event Intervention video reaches 512.3 seconds, spanning both short-horizon observations and extended task sequences.

Across \textsc{P$^3$IBench} we annotate 836 GT response targets. The bottom right of Figure~\ref{fig:composition-duration}  plots these positive GT events by video duration $T$ and normalized trigger time $t_{\mathrm{GT}}/T$, using the start time for interval annotations. As shown, GT events cover the early, middle, and late thirds of videos, accounting for 37.8\%, 39.7\%, and 22.5\% of all events, respectively. Event Intervention is particularly balanced, with 87, 87, and 84 events in these thirds. This distribution provides temporal coverage for evaluating proactive responses at different stages of a video.

\subsection{Evaluation Protocol}
\label{sec:evaluation-protocol}

\textbf{Proactive Task Formulation.} In \textsc{P$^3$IBench}, we formulate proactive video understanding as recurrent decision-making over an incrementally revealed video stream $V$. 
Each sample is initialized with an instruction $I$ that specifies the task scenario and requirements, such as monitoring for the occurrence of a particular event. 
As time progresses, at each sampling step at video time $t$, the model receives the observed frames $V_{\leq t}$ subject to its visual context window, and its response history $H$, then either remains silent or emits a response $o_j=(t_j,\hat{y}_j)$, where $t_j$ is the video time of emission and $\hat{y}_j$ is the response content. 
Given a target response $g_i=(\tau_i,y_i)$, the goal of proactive task evaluation is to assess whether the model can respond accurately and in a timely manner, with $t_j$ approaching the target response time $\tau_i$ and $\hat{y}_j$ approaching the target content $y_i$.


\textbf{Evaluation Metric.}
To jointly capture the accuracy of the response and its temporal alignment, we introduce the \emph{Proactive Response Score (PRS)}. 
For a target
$g_i=(\tau_i,y_i)$ and a model response $o_j=(t_j,\hat{y}_j)$, the pairwise
PRS is calculated as:
\[
s(o_j,g_i)=c(\hat{y}_j,y_i)\exp\left(-\alpha|t_j-\tau_i|\right),
\qquad \alpha=0.05,
\]
where $c(\hat{y}_j, y_i)$ measures the accuracy of the response. For structured outputs, $c\in\{0,1\}$ is determined using deterministic rules, whereas for open-ended responses (i.e., the intervention subset), $c\in[0,1]$ is judged by an LLM. The PRS further applies a temporal decay factor, $\exp\left(-\alpha|t_j-\tau_i|\right)$, which penalizes responses with larger temporal deviations from the target response time.

Considering a single sample may contain multiple target responses, while the model may produce redundant responses or completely miss certain targets, we first match each model output $o_j$ to at most one ground-truth target $g_i$. For each target $g_i$ and its corresponding set of matched responses $O_i$, we aggregate their scores into a target-level score:
\[
S_i=\displaystyle\frac{\max_{o_j\in\mathcal{O}_i}s(o_j,g_i)}{|\mathcal{O}_i|} \;\;\text{if}\;|O_i|>0\;\;\text{else}\;0
\]
 by taking the score of the best-matched response while penalizing redundant responses. The PRS of each sample is subsequently computed as the average target-level score over all targets in the video.

We report two PRS variants w.r.t. different $o_j\rightarrow g_i$ mapping strategies. \textbf{PRS-U} uses unidirectional
matching: each response is assigned to the most recent target at or
before its emission time. Responses emitted before the first target are assigned to that target but receive a score of zero,
regardless of their content; they still count toward the number of
matched responses. \textbf{PRS-B} uses bidirectional matching:
each response is assigned to the temporally nearest target, allowing
early responses to match future targets. Both variants use the same
correctness and aggregation rules.

\section{Experiment}
\subsection{Evaluation Setup}
\label{sec:evaluation-models}

\textbf{Evaluated Models.} For a comprehensive assessment, we evaluate two groups of models: adapted offline MLLMs and native proactive MLLMs.
\textbf{(1) Offline baselines} include GLM-4.1V-9B-Base~\citep{hong2026glm45vglm41vthinkingversatilemultimodal}, InternVL3.5-8B~\citep{wang2025internvl35advancingopensourcemultimodal}, Qwen2.5-VL-3B-Instruct~\citep{bai2025qwen25vltechnicalreport}, Qwen3-VL-4/8B-Instruct~\citep{bai2025qwen3vltechnicalreport}, Qwen3.5-4/9B\citep{qwen3.5} and GPT-5.6-Luna~\citep{openai2026gpt56luna}. These models are originally developed for prompted visual understanding; they are adapted at inference time to process incrementally revealed frames and decide whether to respond. \textbf{(2) Native proactive models} include JoyAI-VL-Interaction~\citep{yao2026joyai}, MMDuet2~\citep{wang2026mmduet2}, Em-Garde~\citep{zheng2026emgarde} and VideoChat3~\citep{li2026videochat3}, which are trained for proactive interaction with streaming video. 

\textbf{Proactive Evaluation Setup.} 
For offline models, we wrap each model with a streaming context management framework to support proactive evaluation. Specifically, (1) a system prompt defines the role, task, and response requirements, including when and how to remain silent; (2) the video is processed in fixed intervals (1 second), with new frames sampled at the default rate of 1 FPS and appended to the input at each timestep, after which the model is prompted to decide whether to respond; and (3) the context retains the initial instruction, as well as the inputs and model outputs from previous rounds. Most models retain all previously observed frames. Due to input context limitations, GPT-5.6-Luna and InternVL3.5-8B retain only the most recent 16 and 32 frames, respectively, while all other textual information is fully preserved. The complete evaluation pipeline is provided in Appendix~\ref{app:streaming_inference}.
For native proactive models, we follow their respective inference frameworks and additionally incorporate the requirements specific to the P3I-Bench tasks into the prompts.

\subsection{Main Results}
\label{sec:main_results}
Table~\ref{tab:main_results} presents model performance across five tasks organized in three categories.
Despite the strong capabilities of MLLMs on many static tasks, their performance remains unsatisfactory in the dynamic proactive setting, with the average PRS scores of most models below 30. The challenge is particularly pronounced for the complex intervention tasks. GPT-5.6-Luna achieves the strongest overall performance, with a notable advantage over open-source models on risk prediction and intervention. However, it still struggles to promptly perceive events and objects as they emerge in the streaming context. Overall, these results suggest that \textbf{the complex contextual dynamics and stringent temporal requirements of proactive tasks remain challenging for current models}. 
Furthermore, we delve deeper to investigate the limitations of existing models based on \textsc{P$^3$IBench}:
\begin{table}[t]
\centering
\caption{\textbf{Model Performance in \textsc{P$^\textbf{3}$IBench}}. PRS-U and PRS-B denote the proactive response score metrics evaluated in unidirectional and bidirectional setups, respectively. In each column, the best performance is highlighted in \textbf{bold} while the runner-up is \underline{underlined}.}
\label{tab:main_results}
\begingroup
\setlength{\tabcolsep}{4pt}
\renewcommand{\arraystretch}{1.15}
\vspace{-4pt}
\resizebox{\textwidth}{!}{%
\begin{tabular}{@{}l|*{2}{c}*{2}{c}|*{2}{c}*{2}{c}|*{2}{c}|*{2}{c}@{}}
\toprule
 & \multicolumn{4}{c|}{\textbf{Perception}} 
 & \multicolumn{4}{c|}{\textbf{Risk Prediction}} 
 & \multicolumn{2}{c|}{\multirow{2}{*}{\textbf{Intervention}}}
 & \multicolumn{2}{c}{\multirow{2}{*}{\textbf{\begin{tabular}[c]{@{}c@{}} Overall\\ Average\end{tabular}}}} \\
\cmidrule{2-9}
\textbf{Model} 
& \multicolumn{2}{c}{\textbf{Event}} 
& \multicolumn{2}{c|}{\textbf{Object}} 
& \multicolumn{2}{c}{\textbf{Accident}} 
& \multicolumn{2}{c|}{\textbf{Crime}} 
& \multicolumn{2}{c|}{} 
& \multicolumn{2}{c}{} \\
& PRS-U & PRS-B & PRS-U & PRS-B 
& PRS-U & PRS-B & PRS-U & PRS-B 
& PRS-U & PRS-B 
& PRS-U & PRS-B \\
\midrule

\multicolumn{13}{l}{\textit{\textbf{Offline Models adapted for Proactive Tasks}}} \\

GLM-4.1V-9B-Base 
& 0.89 & 0.96 & 26.00 & 26.49 
& 2.11 & 2.80 & 18.15 & 19.80 
& 0.93 & 0.93 
& 9.62 & 10.20 \\

InternVL3.5-8B 
& 5.43 & 9.06 & 28.02 & 28.08 
& 23.19 & 38.79 & 43.81 & 52.15 
& 5.89 & 5.89 
& 21.27 & 26.79 \\

Qwen2.5-VL-3B-Instruct 
& 3.44 & 8.27 & 22.38 & 28.44 
& 14.91 & 21.27 & 11.25 & 24.01 
& 1.49 & 1.49 
& 10.69 & 16.70 \\

Qwen3-VL-4B-Instruct 
& 14.84 & 21.35 & 32.67 & 33.82 
& \underline{30.93} & 41.78 & 35.29 & 37.03 
& 9.13 & 9.13 
& 24.57 & 28.62 \\

Qwen3-VL-8B-Instruct 
& \textbf{16.17} & \textbf{26.00} 
& \textbf{44.38} & \textbf{50.33} 
& 29.21 & \underline{68.83} 
& 43.65 & \underline{57.96} 
& 5.53 & 5.53 
& \underline{27.79} & \underline{41.73} \\

Qwen3.5-4B 
& 13.14 & 17.49 & 30.65 & 33.89 
& 18.39 & 27.92 & 36.84 & 46.33 
& 5.17 & 5.17 
& 20.84 & 26.16 \\

Qwen3.5-9B 
& 10.13 & 15.65 & 37.81 & \underline{46.20} 
& 15.82 & 15.82 & 42.75 & 49.62 
& 3.26 & 3.26 
& 21.95 & 26.11 \\

\midrule

GPT-5.6-Luna 
& \underline{15.62} & \underline{24.60} 
& \underline{37.83} & 41.51 
& \textbf{56.41} & \textbf{74.64} 
& \textbf{54.17} & \textbf{62.37} 
& \textbf{29.85} & \textbf{29.85} 
& \textbf{38.78} & \textbf{46.59} \\

\midrule

\multicolumn{13}{l}{\textit{\textbf{Native Proactive Streaming Models}}} \\

Em-Garde 
& 2.00 & 2.41 & 31.92 & 37.61 
& 27.45 & 44.69 & 25.59 & 46.83 
& 4.74 & 8.16 
& 18.34 & 27.94 \\

JoyVL 
& 9.94 & 14.78 & 29.96 & 36.40 
& 19.66 & 50.62 & 43.15 & 52.19 
& 4.10 & 4.10 
& 21.36 & 31.62 \\

MMDuet2 
& 0.18 & 0.26 & 26.00 & 26.00 
& 0.00 & 0.00 & 17.00 & 17.00 
& 1.03 & 1.03 
& 8.84 & 8.86 \\

VideoChat3 
& 6.08 & 12.91 & 25.82 & 27.61 
& 44.09 & 53.38 & 26.83 & 27.90 
& \underline{12.84} & \underline{18.77} 
& 23.13 & 28.11 \\ 


\bottomrule
\end{tabular}%
}
\endgroup
\vspace{-3mm}
\end{table}

\begin{figure}
    \centering
    \includegraphics[width=1.0\linewidth]{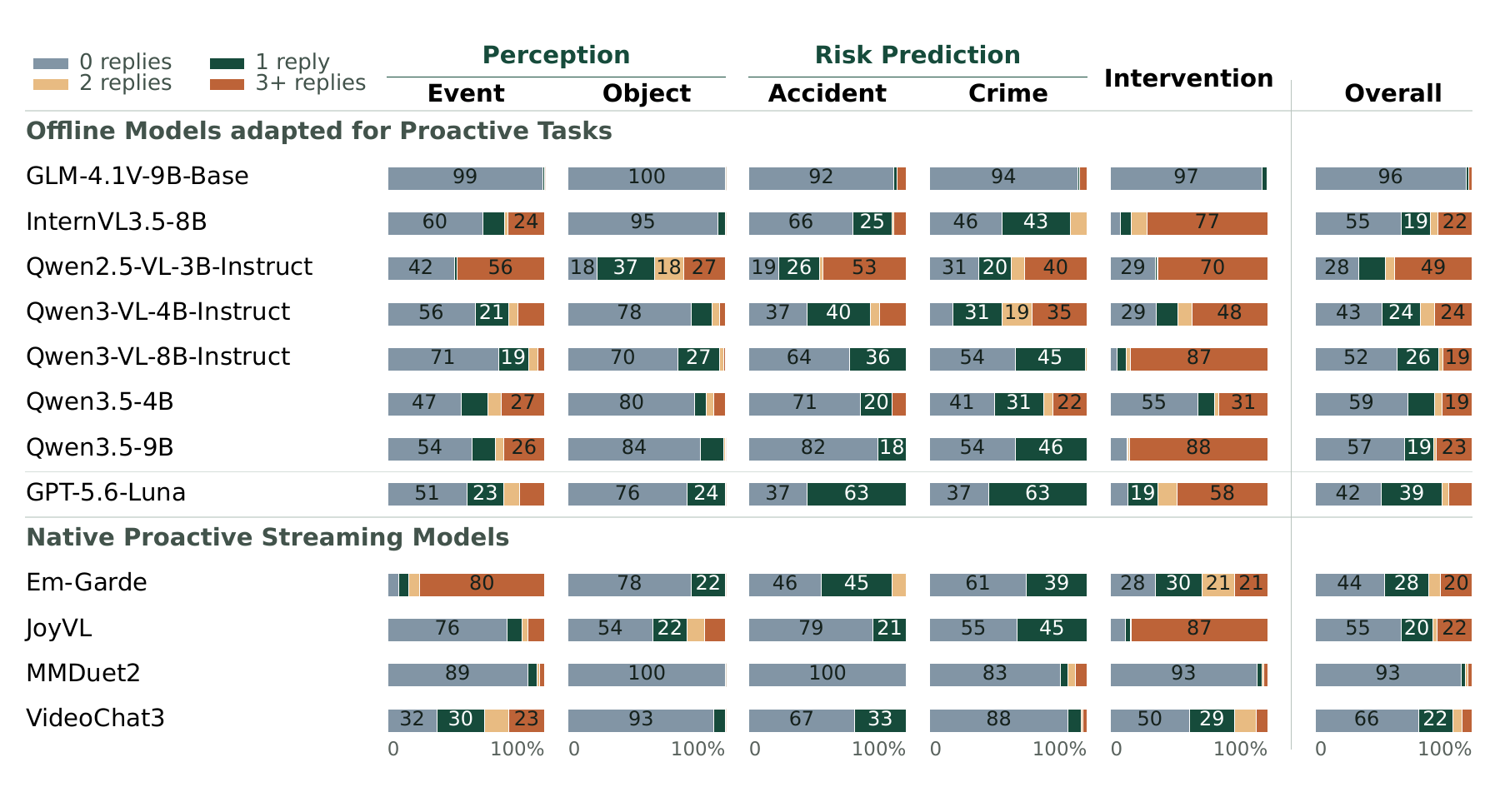}
    \vspace{-10mm}
    \caption{Response frequency distribution of different models across tasks. The response frequency is measured as the number of model responses within the ground-truth window of each target response, grouped into four categories: 0, 1, 2, and $\geq 3$.}
    \label{fig:reply_stats}
    \vspace{-3mm}
\end{figure}

\textbf{(i) Offline models produce excessive, repeated replies.}
Figure~\ref{fig:reply_stats} shows the distribution of the number of responses generated by the models. A major bottleneck for offline models is the difficulty in fully understanding the streaming context and distinguishing between when to remain silent and when to respond. Instead, they tend to continuously generate responses in a manner similar to static QA, resulting in substantial redundancy and consequently incurring penalties in the PRS. 
For example, on the \textit{Accident Warning} task, Qwen2.5-VL-3B produces $\geq3$ responses in 53/100 GT windows with a PRS-B score of 21.27, but would reach 71.39 if the redundancy penalty were removed. Similar behavior is observed across other models, see Appendix~\ref{app:best_reply_results} for further analysis.

\textbf{(ii) Proactive models become overly silent.}
In contrast to offline models, proactive models, which are specifically trained for proactive interaction, tend to reduce redundant responses. However, this may also lead to an over-correction, causing models to remain silent even during intervals when a response is expected. As shown in Figure~\ref{fig:reply_stats}, MMDuet2 and JoyVL miss $79\%$ and $100\%$ of the ground-truth (GT) windows, respectively, on the \textit{Accident Warning} task. We hypothesize that this behavior may stem from the imbalance between the supervision signals for silent responses and content-bearing responses in the training of current proactive models, as discussed in~\citep{xia2026streamingvideoinstructiontuning}. This observation highlights the need to explore more effective proactive training paradigms.

\begin{figure}[t]
    \centering
    \includegraphics[
        width=\linewidth,
        trim={0 11mm 0 18mm},
        clip
    ]{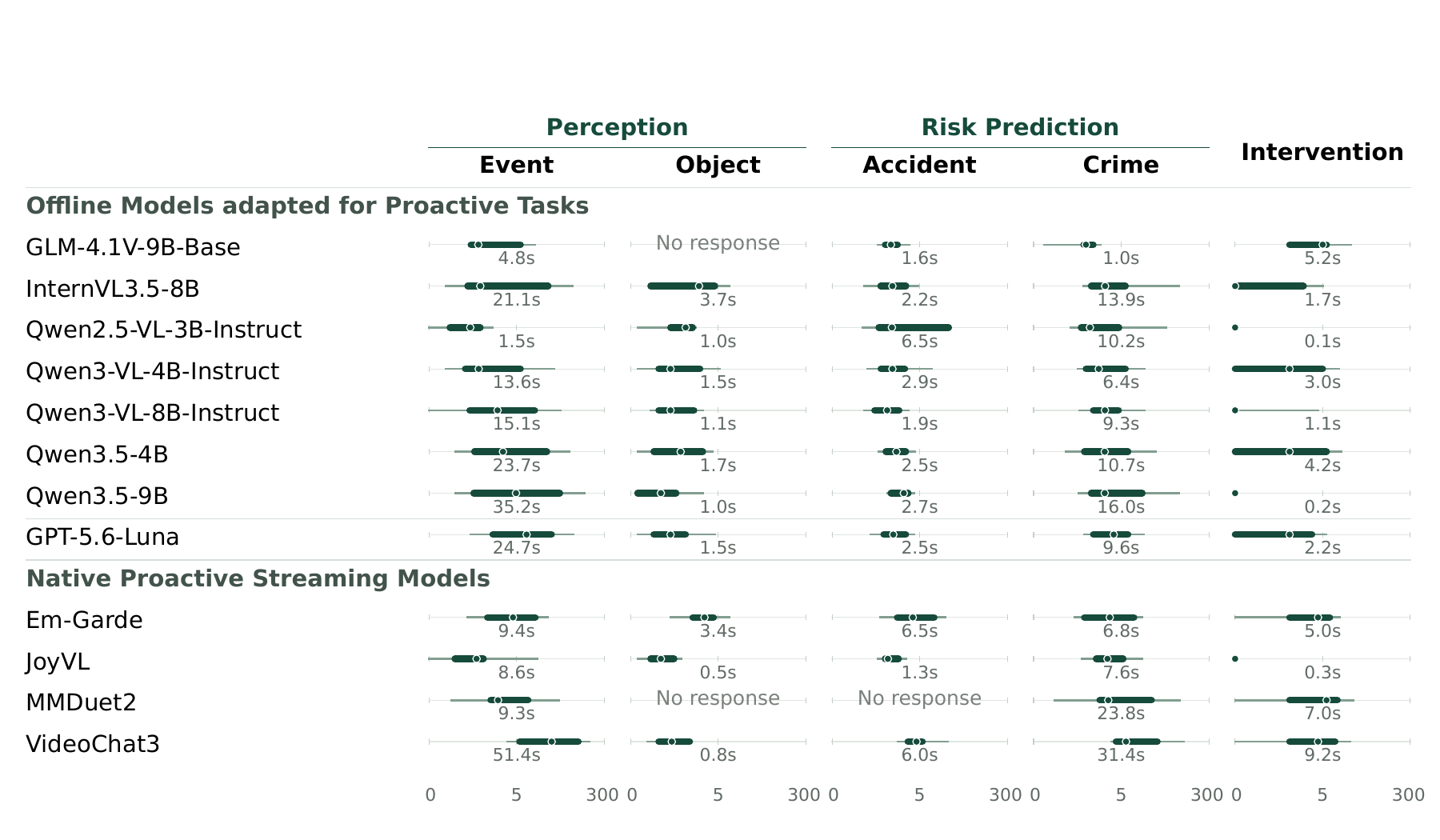}
    \vspace{-4mm}
    \caption{Distribution of the first-response latency across different
    models and tasks. GT windows with no model response are excluded from
    the analysis. Within each distribution, the thin line, thick line,
    and circle denote the 10--90\% percentile range, interquartile range,
    and median, respectively.}
    \label{fig:reply_latency}
\end{figure}



\textbf{(iii) Timeliness remains a major challenge for proactive tasks.}
Beyond response frequency, we find that producing responses at the correct timing remains difficult for current models. 
(1) First, as shown in Figure~\ref{fig:reply_latency}, current models exhibit substantial response latency. 
While their perception of specific objects is relatively timely, their ability to perform event-level perception that requires understanding continuous and complex contexts is substantially weaker. 
This limitation further affects the timeliness of prediction and intervention. 
(2) Second, the comparison between PRS-U and PRS-B in Table~\ref{tab:main_results} reveals a prevalent tendency toward premature responses. 
PRS-B is generally $6$--$15$ points higher than PRS-U, with the gap being particularly pronounced on specific tasks such as \textit{Accident Warning}. 
Although responding in advance can be appropriate in certain scenarios, such as early warning, this discrepancy also suggests that models have difficulty identifying the temporal boundaries of when a response is actually required. 
Therefore, jointly analyzing PRS-B and PRS-U is important for a more comprehensive assessment of the temporal alignment of proactive responses.

In summary, the above quantitative analyses demonstrate that there is a substantial gap between existing models and reliable proactive assistants, as they still need substantial improvements in perceiving dynamic streaming contexts and following complex instructions. In addition, we provide representative examples in Appendix~\ref{app:error_analysis} for a qualitative analysis of common failure modes.

\subsection{Understanding the Challenges behind Streaming Video Contexts}




As discussed in Section~4.2, dynamic context perception remains a key limitation of current models. We therefore further investigate its underlying factors in this section.

\begin{figure}[t]
    \centering
    \includegraphics[
        width=0.9\linewidth,
        trim={0 12mm 0 6mm},
        clip
    ]{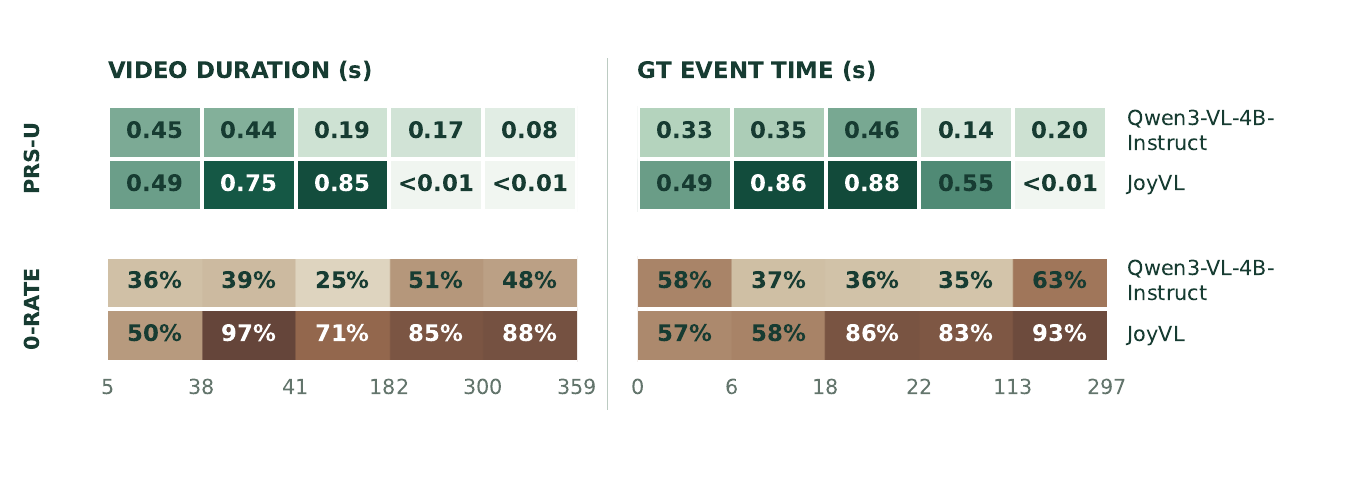}
    \vspace{-4mm}
    \caption{PRS-U of Qwen3-VL-4B-Instruct and JoyVL across different video durations (left) and ground-truth event times (right). Videos are separately binned by each dimension. The top row reports the median PRS-U among non-zero responses, while the bottom row reports the fraction of zero scores. Videos without ground-truth responses are excluded from the video-duration analysis.}
    \label{fig:score-duration-gt-time-heatmap}
    \vspace{-3mm}
\end{figure}

\textbf{Video Length and Event Time.}
As shown in Figure~\ref{fig:score-duration-gt-time-heatmap}, longer videos and events occurring later are more challenging for both models, likely due to the increasing complexity of the accumulated context. In addition, Qwen3-VL is less sensitive to information presented early in the video.



\textbf{Effect of Visual Information Density.}
We further investigate how the density of visual information provided to the model affects its performance. 
(1) We vary the FPS to control the amount of visual information introduced at each query turn. As shown in Table~\ref{tab:fps_comparison}, while a higher FPS can facilitate object perception, it also increases context complexity and makes tasks such as risk prediction more challenging. We therefore retain the default 1-$\mathrm{FPS}$ setting in \textsc{P$^3$IBench}.
(2) We then vary the size of the visual context window to control the density of historical information. In contrast to FPS, the results in Table~\ref{tab:frame_comparison} show that retaining more historical visual information consistently improves performance. Overall, these results suggest that current models are better at processing static historical information, while their ability to model richer dynamic information remains limited.
\begin{table}[t]
\centering
\begingroup
\makeatletter
\renewenvironment{table}[1][]{%
  \begingroup
  \def\@captype{table}%
}{%
  \endgroup
}
\makeatother
\begin{minipage}[t]{0.49\textwidth}
\vspace{0pt}
\begin{table}[t]
\centering
\setlength{\abovecaptionskip}{0pt}
\setlength{\belowcaptionskip}{2pt}
\caption{Effect of the FPS sampling rate.}
\label{tab:fps_comparison}
\begingroup
\fontencoding{T1}\selectfont
\definecolor{mainresultsGreen}{HTML}{164B3B}
\setlength{\tabcolsep}{6pt}
\renewcommand{\arraystretch}{1.4}
\resizebox{0.985\textwidth}{!}{%
\begin{tabular}{@{}lc*{2}{>{\columncolor{mainresultsGreen!9}}c}*{2}{>{\columncolor{mainresultsGreen!24}}c}>{\columncolor{mainresultsGreen!9}}c>{\columncolor{mainresultsGreen!9}[\tabcolsep][0pt]}c@{}}
\hline
 & & \multicolumn{2}{c}{\textbf{Perception}} & \multicolumn{4}{c@{}}{\textbf{Risk Prediction}} \\
\cline{3-4}\cline{5-8}
\textbf{Model} 
& \textbf{FPS} 
& \multicolumn{2}{>{\columncolor{mainresultsGreen!9}}c}{\textbf{Object}} 
& \multicolumn{2}{>{\columncolor{mainresultsGreen!24}}c}{\textbf{Accident}} 
& \multicolumn{2}{>{\columncolor{mainresultsGreen!9}[\tabcolsep][0pt]}c@{}}
{\textbf{Crime}} \\
 & & PRS-U & PRS-B & PRS-U & PRS-B & PRS-U & PRS-B \\
\hline
\multirow{3}{*}{JoyVL} & 0.5 & 29.23 & \textbf{40.76} & 15.09 & 45.35 & \textbf{44.22} & \textbf{53.62} \\
 & 1 & 29.96 & 36.40 & \textbf{19.66} & \textbf{50.62} & 43.15 & 52.19 \\
 & 2 & \textbf{31.50} & 38.21 & 6.54 & 38.78 & 24.05 & 37.07 \\
\hline
\multirow{3}{*}{MMDuet2} & 0.5 & \textbf{26.00} & \textbf{26.00} & \textbf{0.00} & \textbf{0.00} & 16.90 & 16.90 \\
 & 1 & \textbf{26.00} & \textbf{26.00} & \textbf{0.00} & \textbf{0.00} & \textbf{17.00} & \textbf{17.00} \\
 & 2 & \textbf{26.00} & \textbf{26.00} & \textbf{0.00} & \textbf{0.00} & 14.00 & 14.00 \\
\hline
\multirow{3}{*}{\shortstack[l]{Qwen3-VL-\\4B-Instruct}} & 0.5 & 33.46 & 35.01 & 28.76 & 39.86 & 33.83 & 35.43 \\
 & 1 & 32.67 & 33.82 & \textbf{30.93} & \textbf{41.78} & \textbf{35.29} & \textbf{37.03} \\
 & 2 & \textbf{39.26} & \textbf{42.55} & 21.29 & 30.02 & 29.90 & 31.83 \\
\hline
\end{tabular}%
}
\endgroup
\end{table}

\end{minipage}%
\hfill
\begin{minipage}[t]{0.49\textwidth}
\vspace{0pt}
\begin{table}[t]
\centering
\setlength{\abovecaptionskip}{0pt}
\setlength{\belowcaptionskip}{2pt}
\caption{Effect of the limit for historical frames.}
\label{tab:frame_comparison}
\begingroup
\fontencoding{T1}\selectfont
\definecolor{mainresultsGreen}{HTML}{164B3B}
\setlength{\tabcolsep}{6pt}
\renewcommand{\arraystretch}{1.15}
\resizebox{\textwidth}{!}{%
\begin{tabular}{@{}lc*{2}{>{\columncolor{mainresultsGreen!9}}c}*{2}{>{\columncolor{mainresultsGreen!24}}c}>{\columncolor{mainresultsGreen!9}}c>{\columncolor{mainresultsGreen!9}[\tabcolsep][0pt]}c@{}}
\hline
 & & \multicolumn{2}{c}{\textbf{Perception}} & \multicolumn{4}{c@{}}{\textbf{Risk Prediction}} \\
\cline{3-4}\cline{5-8}
\textbf{Model} & \textbf{Frames} & \multicolumn{2}{>{\columncolor{mainresultsGreen!9}}c}{\textbf{Object}} & \multicolumn{2}{>{\columncolor{mainresultsGreen!24}}c}{\textbf{Accident}} & \multicolumn{2}{>{\columncolor{mainresultsGreen!9}[\tabcolsep][0pt]}c@{}}{\textbf{Crime}} \\
 & & PRS-U & PRS-B & PRS-U & PRS-B & PRS-U & PRS-B \\
\hline
\multirow{4}{*}{JoyVL} & 1 & 22.99 & 28.75 & 1.32 & 31.53 & 12.36 & 37.76 \\
 & 2 & 30.49 & 33.87 & 2.88 & 39.07 & 21.38 & 43.51 \\
 & 4 & \textbf{34.24} & \textbf{36.45} & 5.86 & 37.18 & 30.73 & 40.10 \\
 & All & 29.96 & 36.40 & \textbf{19.66} & \textbf{50.62} & \textbf{43.15} & \textbf{52.19} \\
\hline
\multirow{4}{*}{MMDuet2} & 1 & \textbf{26.00} & \textbf{26.00} & \textbf{0.00} & \textbf{0.00} & 17.91 & 17.91 \\
 & 2 & \textbf{26.00} & \textbf{26.00} & \textbf{0.00} & \textbf{0.00} & \textbf{20.13} & \textbf{20.13} \\
 & 4 & \textbf{26.00} & \textbf{26.00} & \textbf{0.00} & \textbf{0.00} & 17.79 & 17.79 \\
 & All & \textbf{26.00} & \textbf{26.00} & \textbf{0.00} & \textbf{0.00} & 17.00 & 17.00 \\
\hline
\multirow{4}{*}{\shortstack[l]{Qwen3-VL-\\4B-Instruct}} & 1 & 25.02 & 25.02 & 12.58 & 20.09 & 28.22 & 29.01 \\
 & 2 & 29.35 & 31.16 & 11.54 & 17.81 & 27.82 & 29.03 \\
 & 4 & 30.15 & \textbf{34.48} & 27.42 & 39.96 & 29.66 & 34.32 \\
 & All & \textbf{32.67} & 33.82 & \textbf{30.93} & \textbf{41.78} & \textbf{35.29} & \textbf{37.03} \\
\hline
\end{tabular}%
}
\endgroup
\end{table}

\end{minipage}
\endgroup
\end{table}

\subsection{Preliminary Exploration to Enhance Proactive Capabilities}



In this section, we conduct a preliminary study on how to address the limitations of existing models discussed above. We hypothesize that a major bottleneck lies in the lack of high-quality training data. Inspired by the high-quality data construction strategy proposed in Section~\ref{sec:task_taxonomy}, we additionally construct a 1.3K training dataset, including 600 Accident Warning samples and 768 Event Perception samples. We fine-tune Qwen3-VL-4B through a two-stage SFT and RL pipeline, together with commonly used Shot2Story~\citep{han2025shot2story} and LLaVA-Video~\citep{zhang2025llavavideo} data. Details of the training pipeline and optimization objectives are provided in Appendix~\ref{app:post-training-details}. Notably, to mitigate the excessive silence observed in Section~\ref{sec:main_results}, we introduce dynamic weighting to balance the supervision strength between different types of responses.

\begin{table}[t]
    \centering
    \setlength{\abovecaptionskip}{0pt}
    \setlength{\belowcaptionskip}{2pt}
    \caption{Effect of post-training on Qwen3-VL-4B-Instruct.}
    \label{tab:model_improvement}

    \footnotesize
    \setlength{\tabcolsep}{3pt}
    \renewcommand{\arraystretch}{0.92}

    \resizebox{1\linewidth}{!}{%
    \begin{tabular}{l|cc|cc|cc|cc|cc|c}
        \toprule
        \multirow{3}{*}{\textbf{Model / Stage}}
        & \multicolumn{4}{c|}{\textbf{Perception}}
        & \multicolumn{4}{c|}{\textbf{Risk Prediction}}
        & \multicolumn{2}{c|}{\multirow{2}{*}{\textbf{Intervention}}}
        & \\
        
        \cmidrule(lr){2-5}
        \cmidrule(lr){6-9}
        
        & \multicolumn{2}{c|}{\textbf{Event}}
        & \multicolumn{2}{c|}{\textbf{Object}}
        & \multicolumn{2}{c|}{\textbf{Accident}}
        & \multicolumn{2}{c|}{\textbf{Crime}}
        & \multicolumn{2}{c|}{}
        & \textbf{PRS-B Avg.} \\
        & PRS-U & PRS-B
        & PRS-U & PRS-B
        & PRS-U & PRS-B
        & PRS-U & PRS-B
        & PRS-U & PRS-B
        & \\
        \midrule
        Qwen3-VL-4B-Instruct
        & \textbf{14.84} & \underline{21.35}
        & \textbf{32.67} & 33.82
        & \underline{30.93} & 41.78
        & \underline{35.29} & 37.03
        & \textbf{9.13} & \textbf{9.13}
        & 28.62 \\

        \quad + SFT
        & 5.84 & 18.10
        & 30.72 & \textbf{48.62}
        & 2.96 & \underline{61.78}
        & 16.84 & 47.04
        & 8.05 & 8.05
        & \underline{36.72} \\

        \quad + RL (P$^3$I-Assistant-4B)
        & 7.47 & \textbf{22.77}
        & \underline{32.35} & \underline{40.03}
        & \textbf{32.88} & \textbf{87.59}
        & 29.67 & \underline{51.21}
        & \underline{9.02} & \underline{9.02}
        & \textbf{42.12} \\
        \midrule
        JoyVL
        & \underline{9.94} & 14.78
        & 29.96 & 36.40
        & 19.66 & 50.62
        & \textbf{43.15} & \textbf{52.19}
        & 4.10 & 4.10
        & 31.62 \\
        \bottomrule
    \end{tabular}%
    }
\vspace{-4mm}
\end{table}


Table~\ref{tab:model_improvement} presents the performance of our fine-tuned model, P$^3$I-Assistant-4B. The results show that well-curated high-quality data can substantially improve proactive capabilities, with both SFT and RL contributing to overall gains of $+8.1$ and $+13.5$ in average PRS-B, respectively. The final model substantially outperforms JoyVL, despite being trained with only 13K samples compared to its 4M training samples, i.e., approximately $300\times$ fewer data. Notably, the gains from post-training also generalize to \textit{Object Perception} and \textit{Crime Alert}, despite no corresponding training data being used. Overall, this preliminary study demonstrates substantial room for further optimization of proactive capabilities, while also highlighting the need for more sophisticated data construction and training strategies for tasks such as \textit{Intervention}, which require complex responses.

\section{Conclusion}

In this work, we introduced \textsc{P$^3$IBench}, a benchmark to evaluate proactive video understanding in real-life streaming scenarios across three major categories: \textbf{Daily Perception}, \textbf{Risk Alert}, and
\textbf{Event Intervention}. Our evaluation of 12 models reveals a persistent gap between video understanding and proactive assistance. To narrow this gap, we train P$^3$I-Assistant-4B through targeted post-training, improving performance on several tasks. We hope that our work will support future research toward more reliable proactive video assistants in dynamic, real-world settings.

\subsection*{AI use statement}
We used generative AI tools to aid and polish writing, draft portions of the manuscript, and assist in generating synthetic or automatically annotated data. All AI-assisted content was reviewed and verified by the authors, who take full responsibility for the final paper.
\subsection*{Ethics statement}
Our benchmark uses existing public research datasets and manually recorded first-person videos. Some tasks involve safety- and crime-related scenarios, and current models may produce incorrect or delayed responses. Therefore, our benchmark is intended for research evaluation rather than deployment in safety-critical or surveillance applications. We follow the licenses and privacy requirements of the corresponding data sources.

\section*{Reproducibility Statement}

We provide detailed descriptions of the construction and evaluation of \textsc{P$^3$IBench} in Section~\ref{sec:task_taxonomy}, and the post-training procedure in Appendix~\ref{app:post-training-details}. Upon publication, we will release the benchmark and training data, data construction pipeline, model checkpoints, training and evaluation code, prompts, and configurations to support reproducibility.





\bibliography{iclr2027_conference}
\bibliographystyle{iclr2027_conference}

\appendix

\section{Task Definitions and Data Examples}
\label{app:task_definitions_examples}

This appendix specifies the tasks introduced in Section~\ref{sec:task_taxonomy} and presents a representative example of each task form. All tasks follow a causal streaming setting: the model receives a task instruction and observes video frames in chronological order, without access to future frames. It should respond when the instructed condition is reliably identified and otherwise remain silent. In the Event Monitoring examples, $C$ denotes the camera wearer. The figures show selected frames with non-uniform temporal spacing; their horizontal distances do not represent elapsed time on a uniform scale. Frames after a ground-truth (GT) response time provide explanatory context and are not available to the model at that response time. Displayed instructions may be shortened for readability.

\subsection{Daily Perception}
\label{app:daily_perception}

\subsubsection{Event Monitoring}
\label{app:event_monitoring}

Event Monitoring uses Ego4D videos and comprises three task forms that differ in the scope of the monitored event and the required response. Each query is provided at the beginning of the video.

\begin{figure}[!htbp]
\setlength{\abovecaptionskip}{0pt}
    \centering
    \includegraphics[width=\linewidth]{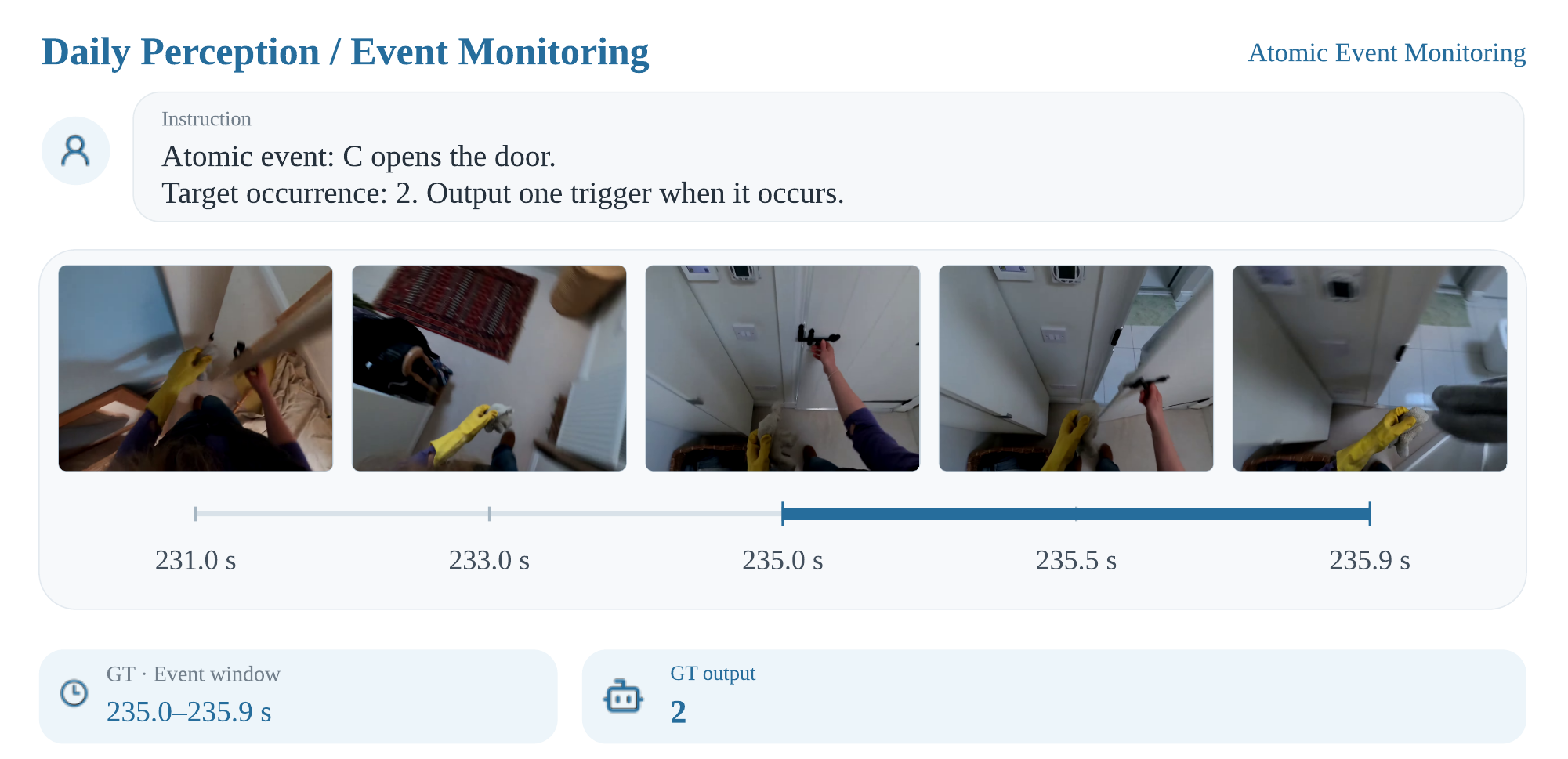}
    \caption{Atomic Event Monitoring. The model is asked to report the second occurrence of opening a door. The selected frames show a local excerpt around the target event; preceding occurrences are outside this excerpt.}
    \label{fig:example_atomic}
\end{figure}

\paragraph{Atomic Event Monitoring.}
An atomic event is a single, concrete action that can be unambiguously identified from visual evidence. Given a query and a target occurrence index $n$, the model counts visually supported occurrences that satisfy all query constraints, treating one continuous action as one occurrence. It must output only the integer $n$, exactly once, as soon as the $n$-th occurrence begins or becomes reliably identifiable, without waiting for its completion. The annotation records the target event's start and end times. Figure~\ref{fig:example_atomic} illustrates this task.

\begin{figure}[!htbp]
\setlength{\abovecaptionskip}{0pt}
    \centering
    \includegraphics[width=\linewidth]{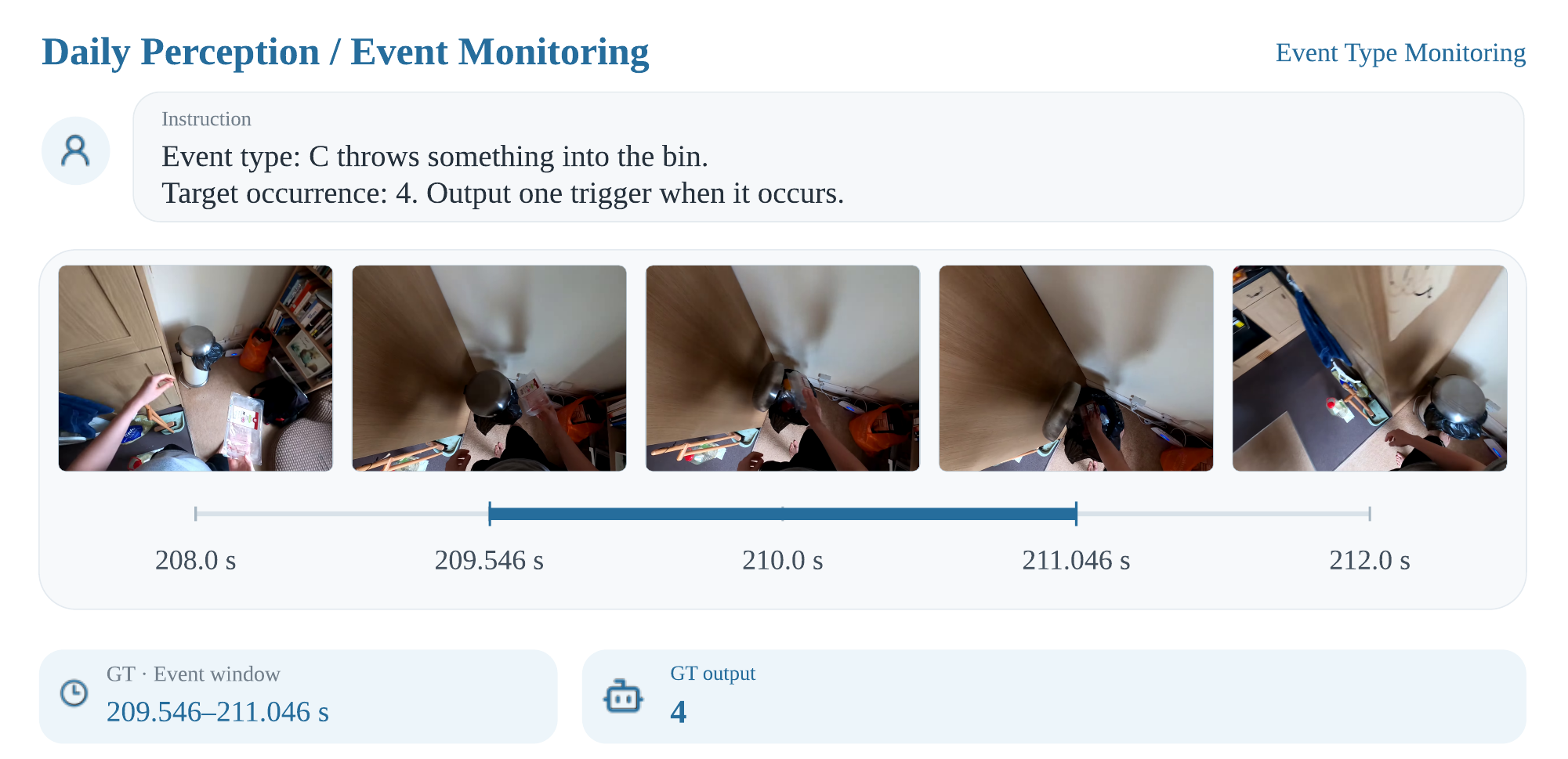}
    \caption{Event Type Monitoring. The query covers throwing something into a bin without restricting the discarded object. The model must output \texttt{4} for the fourth matching occurrence, annotated at 209.546--211.046~s.}
    \label{fig:example_event_type}
\end{figure}

\paragraph{Event Type Monitoring.}
An event type is a well-defined semantic category encompassing related atomic events. Membership depends on their shared event semantics rather than identical objects, tools, locations, manners, or action wording. Given a category query and a target index $n$, the model counts every visually supported occurrence belonging to that category in chronological order. As in Atomic Event Monitoring, the model outputs only $n$, exactly once, when the target occurrence becomes reliably identifiable. The distinction lies in the semantic scope of matching, not the output format. The annotation records the target occurrence's temporal window. Figure~\ref{fig:example_event_type} provides an example.

\begin{figure}[!htbp]
\setlength{\abovecaptionskip}{0pt}
    \centering
    \includegraphics[width=\linewidth]{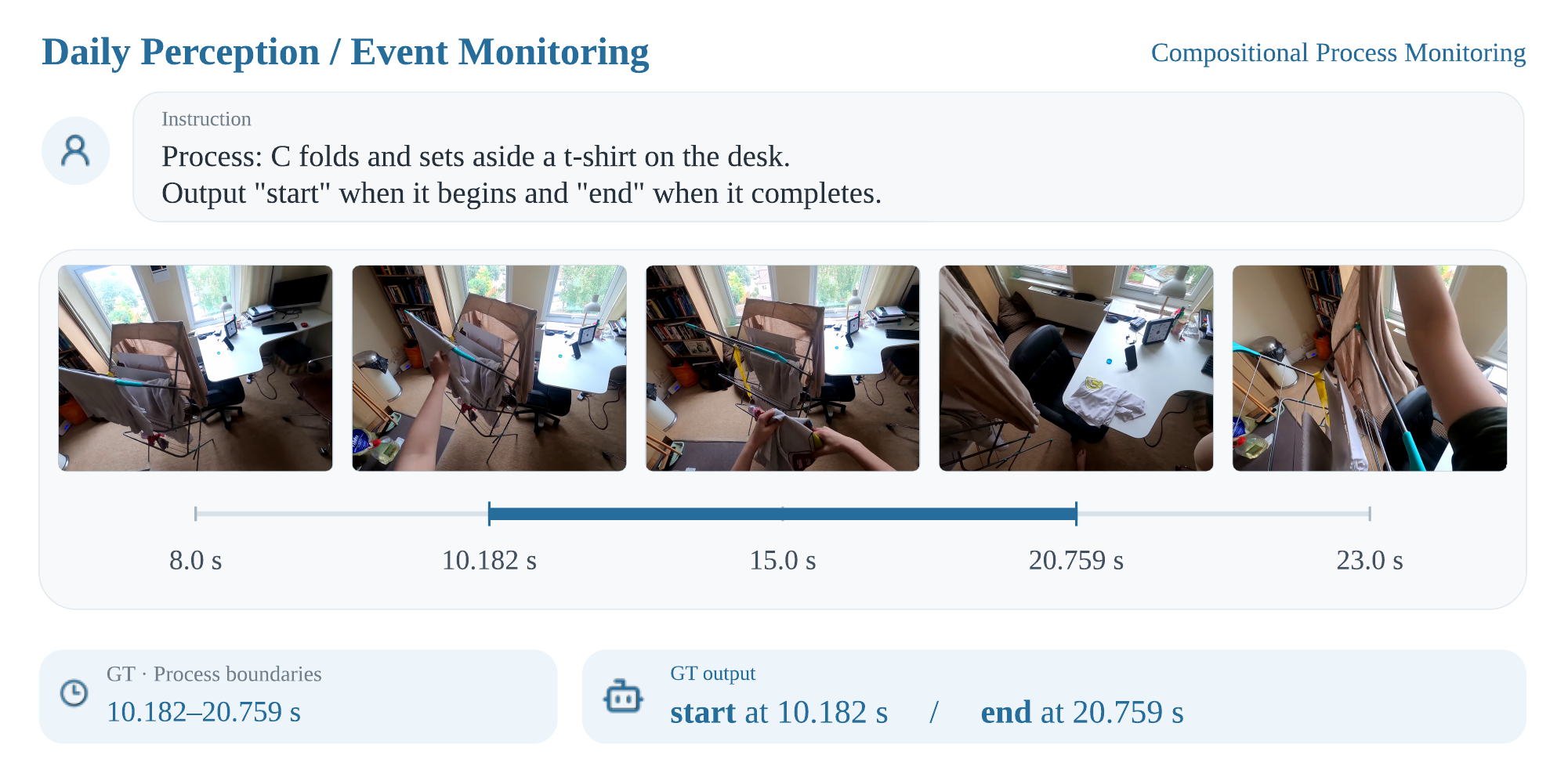}
    \caption{Compositional Process Monitoring. The queried process is folding and setting aside a T-shirt on the desk. Its annotated start and completion are 10.182~s and 20.759~s, respectively. The required responses are \texttt{start} and \texttt{end}.}
    \label{fig:example_compositional}
\end{figure}

\paragraph{Compositional Process Monitoring.}
A compositional process is a complete, coarse-grained activity formed by temporally continuous atomic events that are semantically or functionally related and serve a common goal. Its boundaries follow the complete process specified by the query, excluding independent activities outside that scope. Actions that initiate, advance, adjust, or complete the process remain part of it, even when not explicitly named in the query. The model outputs \texttt{start} when the process begins, followed by \texttt{end} when completion becomes reliably identifiable. Each boundary is reported at most once, and internal actions require no separate response. Figure~\ref{fig:example_compositional} shows the two annotated boundaries.

\subsubsection{Object Monitoring}
\label{app:object_monitoring}

\begin{figure}[!htbp]
\setlength{\abovecaptionskip}{0pt}
    \centering
    \includegraphics[width=\linewidth]{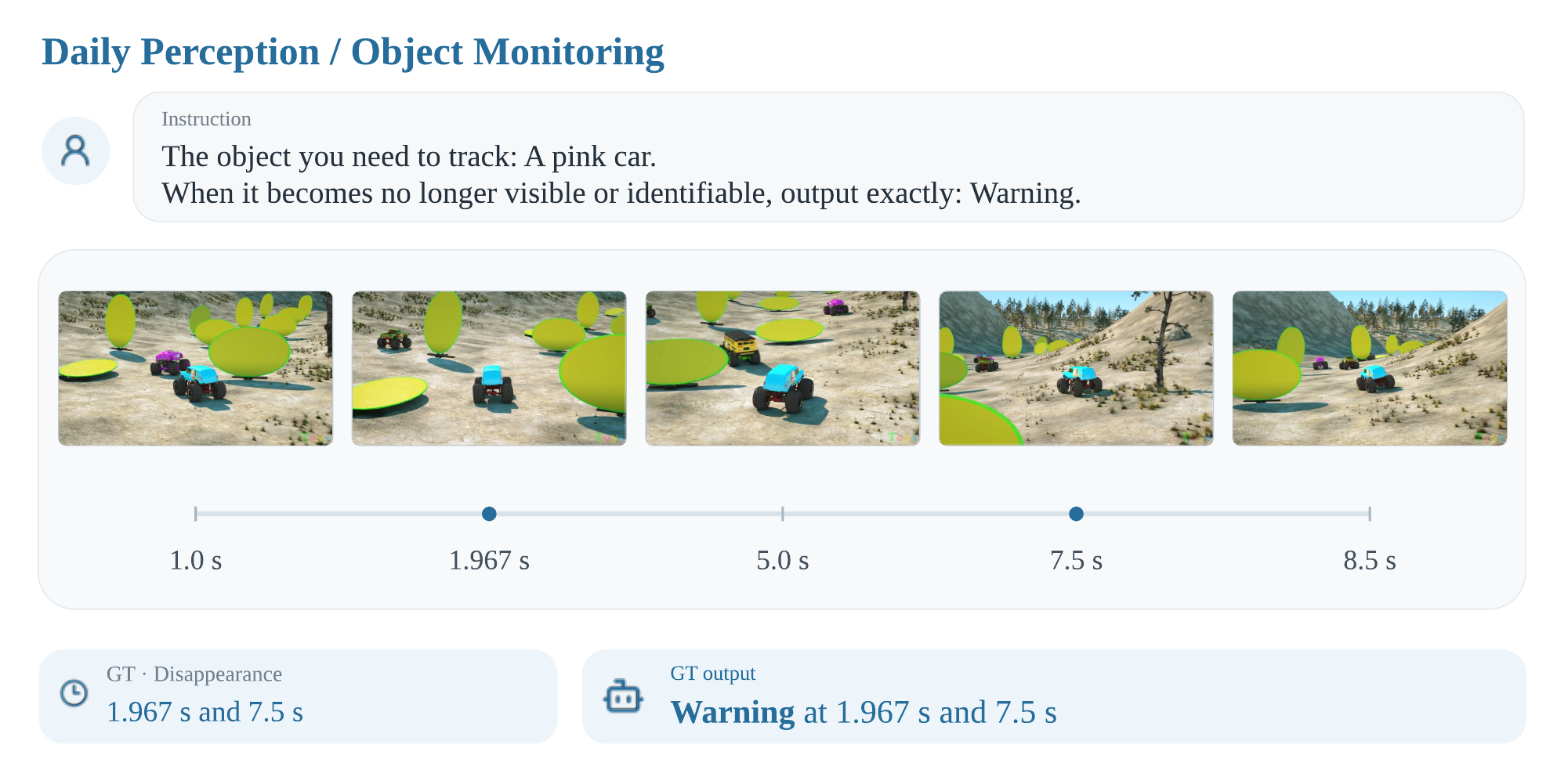}
    \caption{Object Monitoring. The specified object is a pink car. The annotated disappearance times are 1.967~s and 7.5~s, each requiring \texttt{Warning}. The task requires maintaining the target object's identity across the video and distinguishing separate disappearance events.}
    \label{fig:example_object}
\end{figure}

Object Monitoring uses TAG videos and evaluates disappearance detection for an object specified by a natural-language description. The model tracks the object and outputs exactly \texttt{Warning} when it becomes no longer visible or identifiable. Each disappearance requires one response, issued as soon as the disappearance can be reliably confirmed. If the object becomes visible again and subsequently disappears, the later disappearance constitutes a new target; continued absence does not require repeated warnings. The annotations therefore consist of disappearance timestamps paired with the same required warning token. Figure~\ref{fig:example_object} illustrates an instance with two annotated targets.

\FloatBarrier
\subsection{Risk Alert}
\label{app:risk_alert}

\subsubsection{Accident Warning}
\label{app:accident_warning}

\begin{figure}[!htbp]
\setlength{\abovecaptionskip}{0pt}
    \centering
    \includegraphics[width=\linewidth]{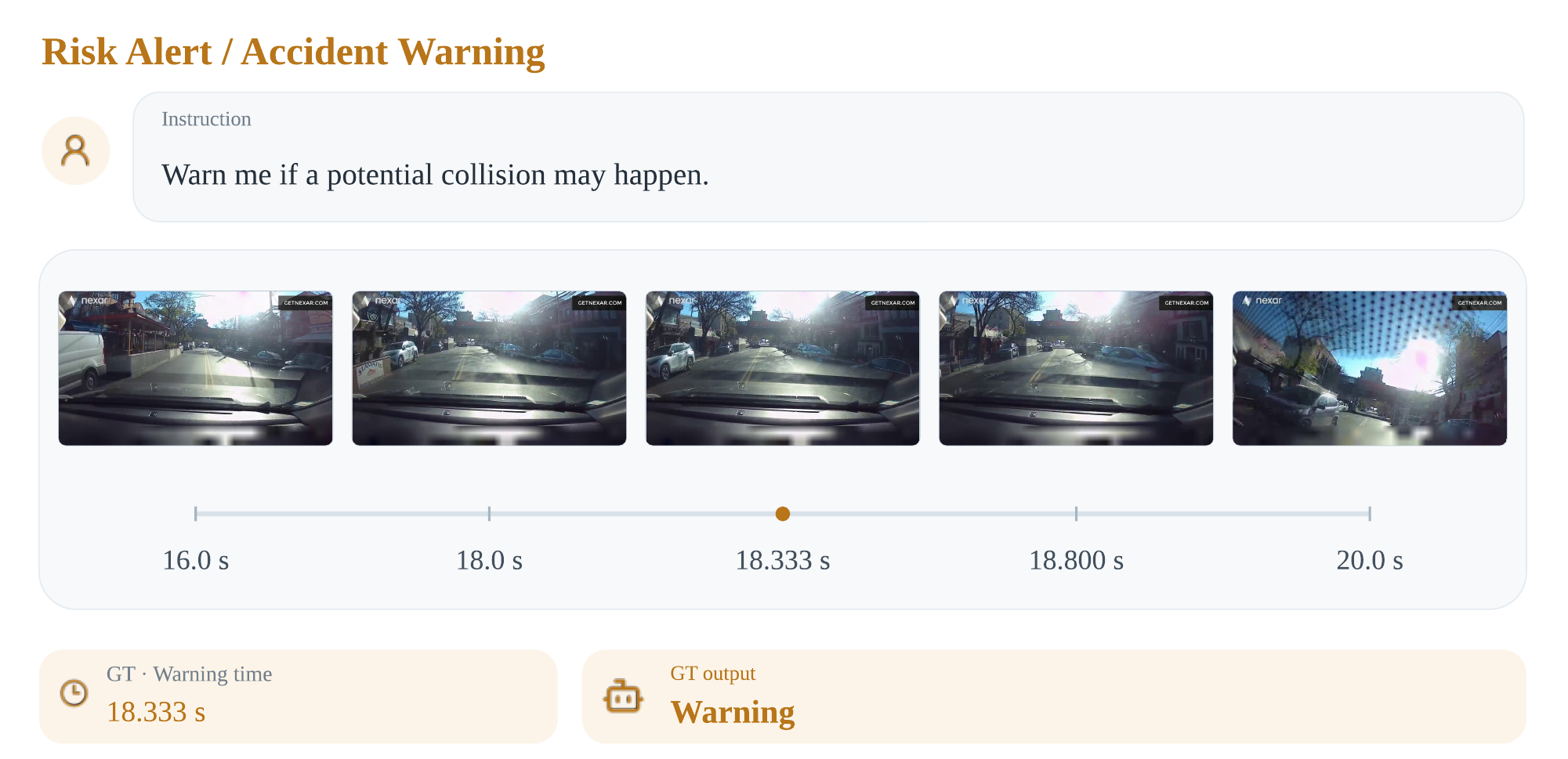}
    \caption{Accident Warning. The model monitors a driving scene for an imminent collision risk, with a GT warning target at 18.333~s. The 18.800~s and 20.0~s frames illustrate subsequent developments and do not provide evidence available at the annotated warning time.}
    \label{fig:example_accident}
\end{figure}

Accident Warning uses Nexar driving videos and requires predicting an imminent collision risk from the stream observed so far. The model outputs exactly \texttt{Warning} at the earliest time when the risk can be reliably identified, before the collision occurs. Only one warning is required per video. Each instance contains a GT warning timestamp, which is distinct from the collision time and serves as the annotated response target. This task concerns anticipatory warning rather than retrospective description of a collision. Figure~\ref{fig:example_accident} shows an example.

\subsubsection{Crime Alert}
\label{app:crime_alert}

\begin{figure}[!htbp]
\setlength{\abovecaptionskip}{0pt}
    \centering
    \includegraphics[width=\linewidth]{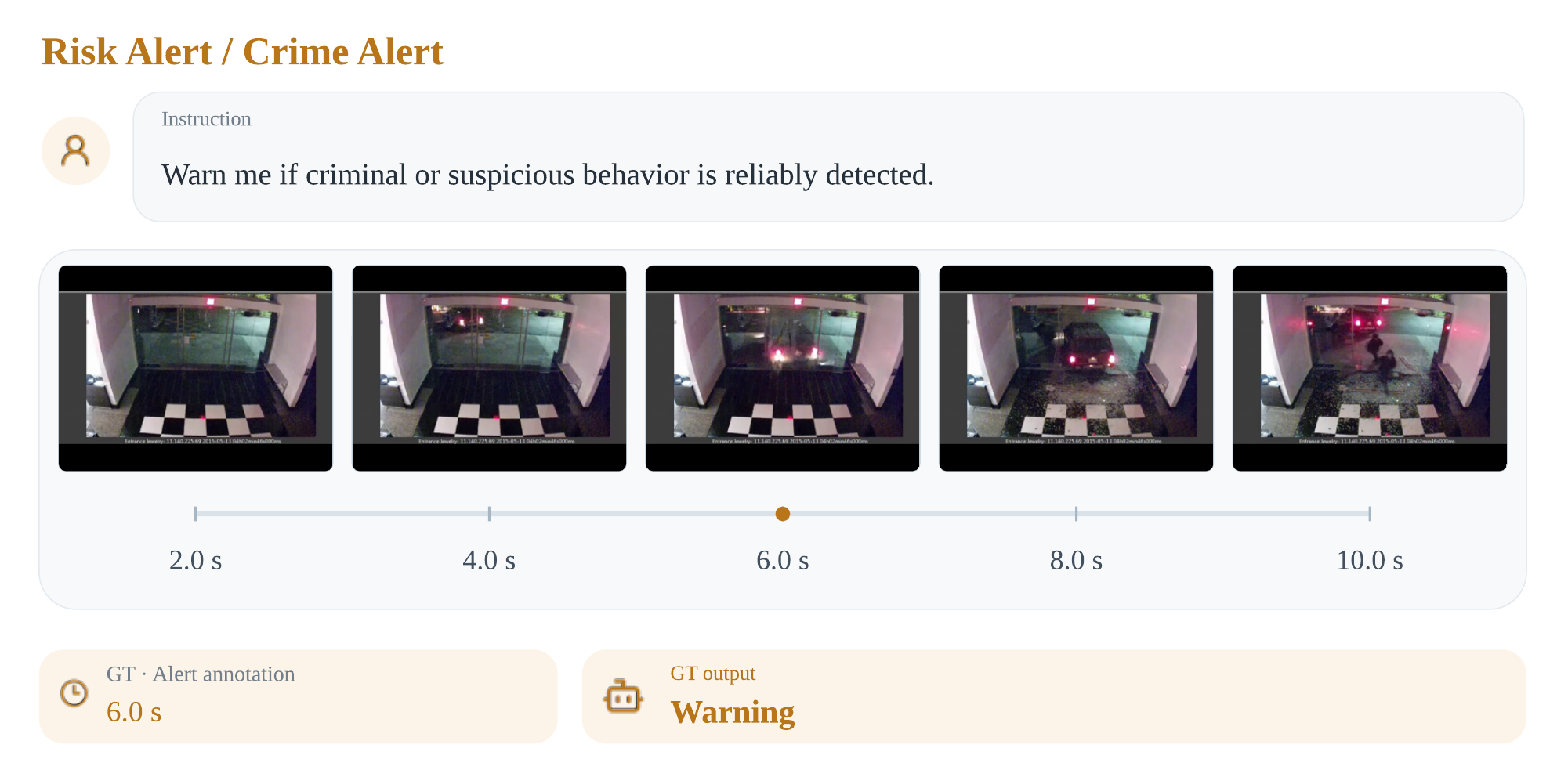}
    \caption{Crime Alert. A surveillance-video example from UCF-Crime has a GT alert target at 6.0~s, with \texttt{Warning} as the required output. The surrounding frames provide context for the developing incident; later frames are not used to justify an earlier response.}
    \label{fig:example_crime}
\end{figure}

Crime Alert uses UCF-Crime videos and requires detecting criminal or suspicious behavior when it becomes visually identifiable. The model outputs exactly \texttt{Warning} at the earliest reliable detection time and only once per video. The annotation supplies the target alert timestamp and the required warning token. Figure~\ref{fig:example_crime} illustrates the task; its displayed instruction summarizes the task-level monitoring requirement.

\FloatBarrier
\subsection{Event Intervention}
\label{app:event_intervention}

Event Intervention uses videos that we manually script and record. Each video is paired with an instruction specifying the user's goal and the conditions under which assistance is expected. The model must relate the observed scene and task progress to these requirements and proactively provide a relevant reminder, correction, or guidance without a follow-up query. Multiple interventions may be needed in one video. Each annotated response contains a start time $t$, a duration $d$, and a natural-language reference answer, defining a response window $[t,t+d]$ during which the intervention remains grounded and useful. Responses should convey the essential information required by the instruction and reference answer, rather than reproduce a fixed string.

\begin{figure}[!htbp]
\setlength{\abovecaptionskip}{0pt}
    \centering
    \includegraphics[width=\linewidth]{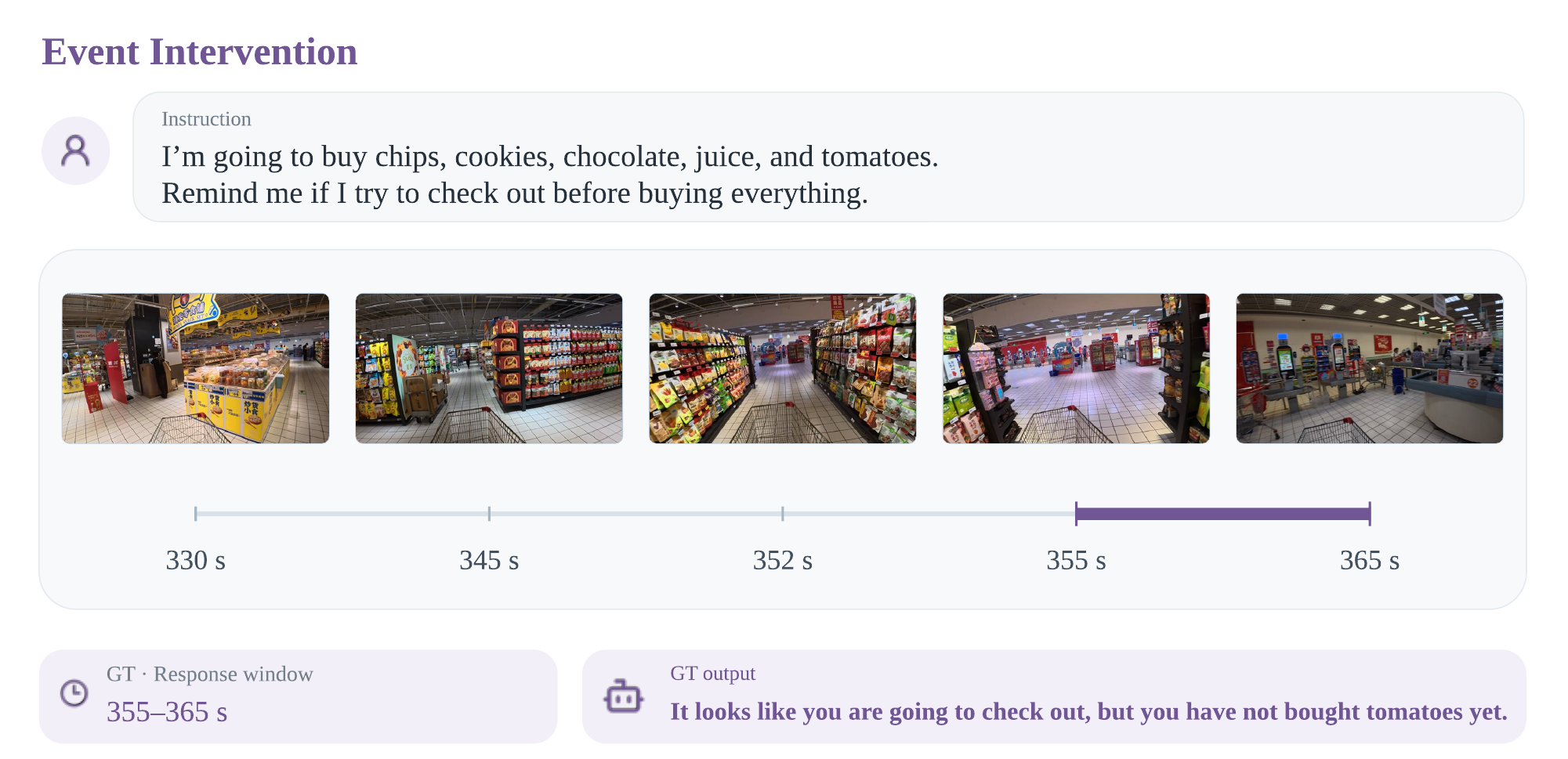}
    \caption{Event Intervention. As the user approaches checkout without having bought tomatoes, the model should remind them of the missing item. The annotated response window is 355--365~s. The reference answer identifies both the attempted checkout and the unfulfilled shopping requirement.}
    \label{fig:example_intervention}
\end{figure}

The shopping example in Figure~\ref{fig:example_intervention} illustrates one intervention within a longer interaction. The full instruction also requests reminders about unpurchased items or their product areas, including relative locations, and corrections when the user picks up an unlisted item. The figure displays an abbreviated instruction focused on checking out before completing the shopping list. The model must retain the outstanding goal across the stream to recognize that tomatoes have not yet been purchased.

\FloatBarrier

\section{Streaming Inference Procedure}
\label{app:streaming_inference}

\paragraph{System prompt.}
The following is the system prompt used for prompt-based inference; \texttt{\{silence\_signal\}} is replaced with the model-specific silence token before inference. A separate user prompt specifies the monitored event or user goal and its response rules.

\definecolor{inferencegreen}{HTML}{205947}
\definecolor{inferencepale}{HTML}{F2F7F4}
\definecolor{inferencerule}{HTML}{B8CEC2}

\begin{lstlisting}[
  basicstyle=\ttfamily\small,
  backgroundcolor=\color{inferencepale},
  frame=single,
  framerule=0.4pt,
  rulecolor=\color{inferencerule},
  framesep=9pt,
  columns=fullflexible,
  keepspaces=true,
  breaklines=true
]
You are a real-time video assistant watching a video frame by frame.
At each turn, you are given newly observed visual frames from the video stream.

Rules:
1. Follow the instruction provided by the user and decide whether to reply.
2. If you do not need to reply now, output exactly "{silence_signal}".

Output only the reply text.
\end{lstlisting}

For most models the silence token is \texttt{Silence}; JoyAI-VL-Interaction uses \texttt{</silence>} and MMDuet2 uses \texttt{NO REPLY}. These tokens are retained in the conversation but removed from the scored responses.

\paragraph{Inference procedure.}
Algorithm~\ref{alg:streaming_inference} formalizes the prompt-based evaluation. The default sampling rate is 1 FPS with one new frame per decision step. Sample times resolving to the same video frame are deduplicated, and time-stamped user instructions join the next eligible frame turn. InternVL3.5-8B and GPT-5.6-Luna retain the latest 32 and 16 images per request, respectively; all other prompt-based models receive every observed image. No visual context limit removes prior instructions or generated replies from the dialogue history.

\newcounter{inferencealgorithm}
\begin{minipage}{\linewidth}
\refstepcounter{inferencealgorithm}\label{alg:streaming_inference}
\noindent\rule{\linewidth}{0.6pt}\par\vspace{0.2em}
\noindent\textbf{Algorithm~\theinferencealgorithm\quad Causal Streaming Inference}\par\vspace{0.2em}
\noindent\rule{\linewidth}{0.4pt}\par\vspace{0.35em}

\noindent\textbf{Require:} Video $V$, instruction $I$, model $M$, sampling rate $f$, frames per turn $B$\par
\noindent\hspace*{4.5em}Scheduled user text $Q$\par
\noindent\textbf{Ensure:} Time-stamped non-silent responses $\mathcal{O}$
\vspace{0.45em}

\noindent\begin{tabularx}{\linewidth}{@{}r@{\hspace{0.65em}}>{\raggedright\arraybackslash}X@{}}
1: & $s \leftarrow \mathrm{SilenceSignal}(M)$ \\
2: & $H \leftarrow [\mathrm{SystemPrompt}(M,s),\,\mathrm{TaskPrompt}(I,s)]$ \\
3: & $\mathcal{O} \leftarrow \emptyset$ \\
4: & $F \leftarrow \mathrm{DecodeUniqueFrames}(V,f)$ \hfill {\scriptsize$\triangleright$ Chronological samples} \\
5: & \textbf{for} each consecutive group $G$ of $B$ frames in $F$ \textbf{do} \\
6: & \hspace*{1.3em}$t \leftarrow \mathrm{time}(\mathrm{last}(G))$ \\
7: & \hspace*{1.3em}$q \leftarrow \mathrm{DueText}(Q,t)$ \\
8: & \hspace*{1.3em}$H \leftarrow H \mathbin{\|} [\mathrm{UserTurn}(G,q,t)]$ \\
9: & \hspace*{1.3em}$r \leftarrow \mathrm{Inference}(M,H)$ \\
10: & \hspace*{1.3em}$H \leftarrow H \mathbin{\|} [\mathrm{AssistantTurn}(r,t)]$ \\
11: & \hspace*{1.3em}\textbf{if} $r \ne s$ \textbf{then} \\
12: & \hspace*{2.6em}$\mathcal{O} \leftarrow \mathcal{O} \mathbin{\|} [(t,r)]$ \\
13: & \hspace*{1.3em}\textbf{end if} \\
14: & \textbf{end for} \\
15: & \textbf{return} $\mathcal{O}$ \\
\end{tabularx}
\vspace{0.15em}\noindent\rule{\linewidth}{0.6pt}
\end{minipage}

\section{Training Details of \texorpdfstring{P$^3$I}{P3I}-Assistant-4B}
\label{app:post-training-details}

\subsection{Training Data}

The SFT data consist of 12,000 broadly scoped proactive video question-answering examples and 1,368 additionally annotated examples targeting the tasks in our benchmark. The broadly scoped data include 10,000 PVQ examples derived from Shot2Story and 2,000 CRR examples from LLaVA-Video. The additionally annotated data comprise 600 Accident examples and 768 Event examples, where Event collectively denotes the Atomic, Event Type, and Compositional tasks, containing 259, 232, and 277 examples, respectively. All examples are combined into a deterministically shuffled training file without additional task-dependent resampling.

The RL data contain 863 rollout windows, including 300 Accident windows and 563 Event windows. The Event portion consists of 100 Atomic, 100 Event Type, and 363 Compositional windows. For Compositional, the start and end targets of 191 videos can be covered by the same rollout window, whereas the remaining 86 videos are divided into separate start- and end-oriented windows. This produces \(191+86\times2=363\) rollout windows from 277 source videos.

\subsection{Supervised Fine-Tuning}

The SFT stage is initialized from Qwen3-VL-4B-Instruct. The vision encoder is frozen, while the language model and visual merging and alignment modules remain trainable. Each training example retains the preceding visual observations and alternating interaction history within the sampled video clip. The target at each interaction turn is either \texttt{</Silence>} or a response beginning with \texttt{</Response>}, followed by a required signal or a natural-language answer. Each Compositional example contains two boundary responses corresponding to its start and end targets.

To account for the imbalance between silence and response decisions, focal and frequency-balancing factors are applied to the two state tokens. Let \(\mathcal{V}\) denote the set of supervised token positions, excluding positions whose labels are masked with \(-100\). The cross-entropy loss at position \(i\) is

\[
\ell_i=-\log p_{\theta}(y_i\mid x,y_{<i}),
\]

where \(x\) denotes the video, task instruction, and interaction history; \(y_i\) is the target token at position \(i\); \(y_{<i}\) denotes the preceding target tokens; and \(\theta\) denotes the trainable model parameters. Let \(\mathcal{S}\subseteq\mathcal{V}\) denote positions whose targets are \texttt{</Silence>} or \texttt{</Response>}. The weight assigned to each supervised token is

\[
w_i=
\begin{cases}
\displaystyle
\frac{N_s}{K n_{y_i}}
\left(1-\exp(-\ell_i)\right)^{\gamma},
& i\in\mathcal{S},\\[8pt]
1, & i\notin\mathcal{S},
\end{cases}
\]

where \(n_{y_i}\) is the number of occurrences of state token \(y_i\) in the current batch, \(N_s\) is the total number of supervised state tokens in that batch, \(K\) is the number of state-token classes present in the batch, and \(\gamma\) is the focal exponent. The final SFT loss is

\[
\mathcal{L}_{\mathrm{SFT}}
=
\frac{1}{|\mathcal{V}|}
\sum_{i\in\mathcal{V}}w_i\ell_i,
\]

where \(|\mathcal{V}|\) is the total number of supervised tokens. Ordinary response tokens therefore retain the standard cross-entropy loss, while difficult and less frequent state-token decisions receive larger weights. All supervised tokens share the same token-count normalization, without separately weighting state-token and response-content losses.

The SFT stage takes approximately 2 hours and 49 minutes to complete. The remaining training configuration is summarized in Table~\ref{tab:sft_configuration}.

\begin{table*}[t]
    \centering
    \caption{\textbf{Configuration for supervised fine-tuning.}}
    \label{tab:sft_configuration}
    \small
    \setlength{\tabcolsep}{8pt}
    \begin{tabular}{p{0.28\textwidth}p{0.64\textwidth}}
        \toprule
        \textbf{Configuration} & \textbf{Value} \\
        \midrule
        Initialization & Qwen3-VL-4B-Instruct \\
        Video sampling rate & 1 FPS \\
        Learning rate & \(1\times10^{-5}\) \\
        Training epochs / steps & 1 / 418 \\
        GPUs & 4 \\
        Per-device batch size & 1 \\
        Gradient accumulation steps & 8 \\
        Optimizer & Fused AdamW \\
        Learning-rate schedule & Cosine decay \\
        Warmup ratio & 5\% \\
        \bottomrule
    \end{tabular}
\end{table*}

\subsection{Reinforcement Learning}

The RL stage is initialized from the final SFT checkpoint and optimized using Group Relative Policy Optimization (GRPO). Rollouts reveal the video incrementally by introducing newly sampled frames over successive interaction rounds. Accident and Compositional rollouts retain a bounded video prefix preceding the rollout window. For Atomic and Event Type, the historical context begins at the start of the video, allowing the model to determine how many times the queried event has occurred. Responses preceding the rollout window are treated as a fixed prefix, and only responses generated within the rollout participate in policy optimization. The complete RL configuration is reported in Table~\ref{tab:rl_configuration}.

\begin{table*}[!t]
    \centering
    \caption{\textbf{Configuration for reinforcement learning.}}
    \label{tab:rl_configuration}
    \small
    \setlength{\tabcolsep}{8pt}
    \begin{tabular}{p{0.30\textwidth}p{0.62\textwidth}}
        \toprule
        \textbf{Configuration} & \textbf{Value} \\
        \midrule
        Initialization & Final SFT checkpoint at step 418 \\
        Optimization algorithm & Group Relative Policy Optimization (GRPO) \\
        Learning rate & \(5\times10^{-7}\) \\
        Training epochs / steps & Approximately 1 / 143 \\
        Training GPUs / rollout GPUs & 3 / 1 \\
        Per-device batch size & 1 \\
        Gradient accumulation steps & 8 \\
        Generations per input & 4 \\
        GRPO clipping coefficient & 0.2 \\
        KL coefficient & 0 \\
        Optimizer & Fused AdamW \\
        Learning-rate schedule & Cosine decay \\
        Warmup ratio & 5\% \\
        Generation temperature & 0.9 \\
        Top-\(p\) / top-\(k\) & 0.9 / 50 \\
        Video sampling rate & 1 FPS \\
        Frames per interaction turn & 1 \\
        \bottomrule
    \end{tabular}
\end{table*}

The reward is computed over the complete generated trajectory. The temporal score assigned to a response emitted at time $t$ is

\[
s(t)=\exp\left(-\alpha\left|t-t_{\mathrm{GT}}\right|\right),
\]

where $t$ is the response time, $t_{\mathrm{GT}}$ is the annotated target time, and $\alpha$ controls the decay with respect to temporal distance. Following the benchmark evaluation protocol, we set $\alpha=0.05$.

For Accident, let \(\mathcal{J}\) denote the set of non-silent outputs whose normalized content strictly matches \texttt{Warning}, and let \(M\) denote the total number of non-silent outputs, including malformed outputs. If at least one correct warning is generated, the closest one is selected as

\[
j^{*}
=
\arg\min_{j\in\mathcal{J}}
\left|t_j-t_{\mathrm{GT}}\right|,
\]

where \(t_j\) is the emission time of output \(j\). The Accident reward is

\[
R_{\mathrm{Accident}}
=
\begin{cases}
s(t_{j^{*}})-\lambda(M-1), & |\mathcal{J}|>0,\\
-\lambda M, & |\mathcal{J}|=0,
\end{cases}
\]

where \(\lambda=0.2\) is the penalty assigned to each unmatched non-silent output. When at least one correct warning is produced, the closest correct warning contributes its temporal score, while the remaining outputs are penalized. If no correct warning is produced, all non-silent outputs are penalized.

For Atomic and Event Type, a response is considered correct only when its predicted integer exactly matches the target. Their trajectory reward is

\[
R_{\mathrm{Event}}
=
\operatorname{PRS\text{-}B}_{\mathrm{first}}
\left(\mathcal{O},\mathcal{G}\right),
\]

where \(\mathcal{O}\) is the set of non-silent outputs in the rollout, \(\mathcal{G}\) is the set of ground-truth event responses, and \(\operatorname{PRS\text{-}B}_{\mathrm{first}}\) denotes the bidirectional first-only score used in our evaluation. It selects the temporally closest semantically correct output while retaining all non-silent outputs in the score normalization. Incorrect and repeated responses therefore reduce the trajectory reward.

For Compositional, predictions are first assigned to the \texttt{start} and \texttt{end} labels and then matched strictly against their corresponding targets. Let \(\mathcal{A}\) denote the set of active boundary labels in the current rollout, \(S_l\) denote the bidirectional first-only score for label \(l\), and \(N_{\mathrm{inv}}\) denote the number of malformed predictions or predictions associated with a boundary absent from the current rollout. The reward is

\[
R_{\mathrm{Comp}}
=
\frac{1}{|\mathcal{A}|}
\sum_{l\in\mathcal{A}}S_l
-\lambda N_{\mathrm{inv}},
\]

where \(|\mathcal{A}|\) is the number of active boundary labels and \(\lambda=0.2\) is the invalid-output penalty. Repeated predictions of an active boundary remain in the corresponding score denominator and consequently reduce \(S_l\). A valid \texttt{</Silence>} output is excluded from response scoring and receives no independent positive reward. Consequently, remaining silent throughout a rollout does not yield the temporal reward associated with its target responses.

The explicit additive penalties in the Accident and Compositional rewards are introduced because the SFT checkpoint exhibits a stronger tendency to generate redundant or invalid outputs on these two tasks. No analogous additive penalty is applied to the remaining task subsets, but incorrect or repeated outputs still reduce their metric-based trajectory rewards.
\section{Error Analysis}
\label{app:error_analysis}

\begin{figure}[t]
\setlength{\abovecaptionskip}{0pt}
    \centering
    \includegraphics[width=\linewidth]{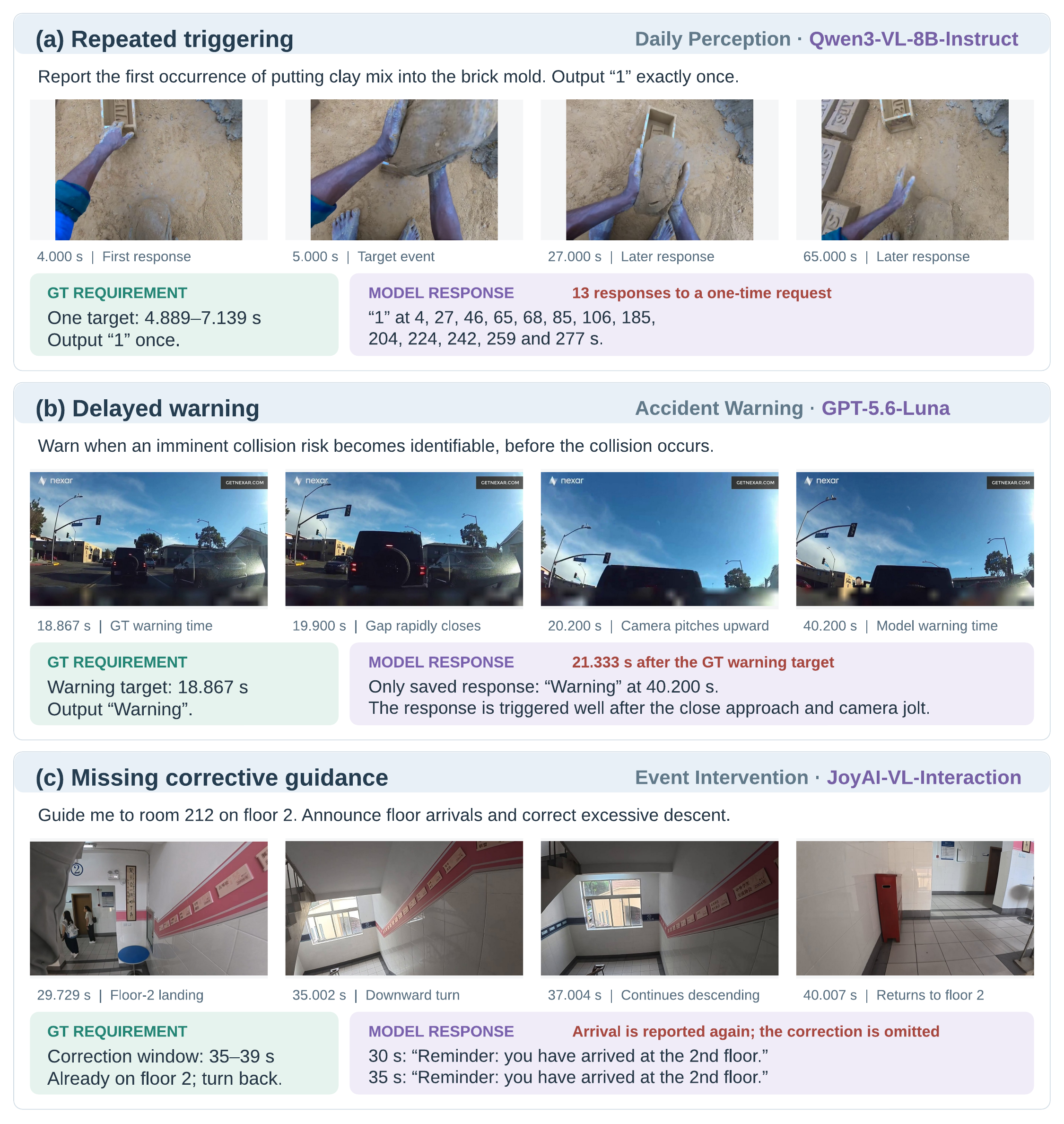}
    \caption{Failure cases from model response traces. (a) Repeated outputs for a single requested event occurrence. (b) A warning recorded 21.333\,s after the annotated warning target. (c) A repeated floor-arrival notice that omits the required correction. Instructions and GT descriptions are abridged. Later frames provide context and do not justify earlier responses.}
    \label{fig:error_analysis}
\end{figure}

Figure~\ref{fig:error_analysis} presents three failure cases from model response traces, illustrating distinct limitations in response control, warning timeliness, and goal-conditioned intervention.

\paragraph{Repeated triggering.}
In Figure~\ref{fig:error_analysis}(a), the instruction requires a single response for the first occurrence of placing clay mix into a brick mold. Qwen3-VL-8B-Instruct nevertheless outputs \texttt{1} at 13 timestamps between 4 and 277\,s. Subsequent molding actions do not constitute additional requested targets. Although the first response slightly precedes the annotated onset, the repeated outputs violate the explicit one-response constraint. This case illustrates a failure to maintain selective responding after a one-time request has been addressed.

\paragraph{Delayed warning.}
In Figure~\ref{fig:error_analysis}(b), the GT warning target is 18.867\,s. The video subsequently shows the gap to the preceding vehicle closing rapidly, followed by a sharp upward camera pitch around 20.2\,s. The only recorded \texttt{Warning} from GPT-5.6-Luna occurs at 40.200\,s, near the video end and 21.333\,s after the GT target. The warning therefore follows the close approach and abrupt camera motion instead of anticipating them. This trace illustrates how a valid warning token can arrive too late to serve its intended proactive role.

\paragraph{Missing corrective guidance.}
In Figure~\ref{fig:error_analysis}(c), the user requests guidance to room 212 on the second floor, including corrections for excessive descent. JoyAI-VL-Interaction correctly announces arrival at the second floor at 30\,s. The user then turns toward the downward staircase and begins descending. During the annotated correction window of 35--39\,s, the model only repeats the floor-arrival notice at 35\,s, without advising the user to stop or turn back. Recognizing the floor is therefore insufficient: useful intervention requires identifying deviations of the unfolding action from the intended destination and providing the appropriate correction.

\section{Best-Reply Evaluation without Redundancy Penalty}
\label{app:best_reply_results}

Table~\ref{tab:best_reply_results} reports an alternative to the PRS results in Table~\ref{tab:main_results}. It retains the same response-to-target assignment, correctness score, and temporal decay as PRS-U and PRS-B, but measures whether a model produces \emph{at least one} well-timed, correct response to each target. For the responses $\mathcal{O}_i$ matched to target $g_i$, the target-level score is
\[
S_i^{\mathrm{best}}=
\begin{cases}
\displaystyle\max_{o_j\in\mathcal{O}_i}s(o_j,g_i), & |\mathcal{O}_i|>0,\\
0, & |\mathcal{O}_i|=0.
\end{cases}
\]
Unlike the primary PRS, this variant does not divide by the number of matched responses. Targets requiring silence retain the original scoring rule. Target scores are averaged within each video and then across videos; the Event column averages the Atomic, Compositional, and Event Type subsets, and Overall averages the five tasks equally. The results are reported on a 0--100 scale.

\begin{table}[t]
\centering
\caption{\textbf{Best-reply model performance in \textsc{P$^\textbf{3}$IBench}.} PRS-U' and PRS-B' retain the original matching and temporal decay rules but score each target by its highest-scoring matched reply, without dividing by the number of matched replies. The best and second-best available scores in each column are \textbf{bold} and \underline{underlined}, respectively.}
\label{tab:best_reply_results}
\begingroup
\setlength{\tabcolsep}{4pt}
\renewcommand{\arraystretch}{1.15}
\resizebox{\textwidth}{!}{%
\begin{tabular}{@{}l|*{2}{c}*{2}{c}|*{2}{c}*{2}{c}|*{2}{c}|*{2}{c}@{}}
\toprule
 & \multicolumn{4}{c|}{\textbf{Perception}}
 & \multicolumn{4}{c|}{\textbf{Risk Prediction}}
 & \multicolumn{2}{c|}{\multirow{2}{*}{\textbf{Intervention}}}
 & \multicolumn{2}{c}{\multirow{2}{*}{\textbf{\begin{tabular}[c]{@{}c@{}} Overall\\ Average\end{tabular}}}} \\
\cmidrule{2-9}
\textbf{Model}
& \multicolumn{2}{c}{\textbf{Event}}
& \multicolumn{2}{c|}{\textbf{Object}}
& \multicolumn{2}{c}{\textbf{Accident}}
& \multicolumn{2}{c|}{\textbf{Crime}}
& \multicolumn{2}{c|}{}
& \multicolumn{2}{c}{} \\
& PRS-U' & PRS-B' & PRS-U' & PRS-B'
& PRS-U' & PRS-B' & PRS-U' & PRS-B'
& PRS-U' & PRS-B'
& PRS-U' & PRS-B' \\
\midrule
\multicolumn{13}{l}{\textit{\textbf{Offline Models adapted for Proactive Tasks}}} \\
GLM-4.1V-9B-Base & 0.89 & 0.96 & 26.00 & 26.49 & 7.38 & 8.08 & 21.76 & 23.41 & 1.32 & 1.32 & 11.47 & 12.05 \\
InternVL3.5-8B & 28.06 & 33.03 & 28.02 & 28.08 & 30.71 & 46.83 & 48.10 & 58.63 & 37.06 & 37.06 & 34.39 & 40.73 \\
Qwen2.5-VL-3B-Instruct & 31.70 & 38.52 & \textbf{63.20} & \textbf{69.27} & \textbf{63.74} & 71.39 & 46.25 & \underline{60.82} & 16.98 & 16.98 & 44.37 & 51.40 \\
Qwen3-VL-4B-Instruct & \underline{34.22} & \underline{43.97} & 40.32 & 42.41 & 55.56 & 69.23 & \textbf{69.72} & \textbf{74.28} & 31.00 & 31.00 & \textbf{46.16} & \underline{52.18} \\
Qwen3-VL-8B-Instruct & 20.77 & 31.33 & \underline{48.20} & 53.90 & 33.06 & \underline{72.67} & 44.51 & 59.30 & \underline{39.88} & \underline{39.88} & 37.28 & 51.42 \\
Qwen3.5-4B & 29.74 & 36.78 & 33.34 & 36.57 & 25.47 & 35.00 & 48.17 & 57.94 & 14.51 & 14.51 & 30.24 & 36.16 \\
Qwen3.5-9B & 25.26 & 31.86 & 39.12 & 47.20 & 15.82 & 15.82 & 43.20 & 50.06 & 35.82 & 35.82 & 31.84 & 36.15 \\
\midrule
GPT-5.6-Luna & 26.04 & 38.13 & 39.82 & 43.49 & \underline{56.41} & \textbf{74.64} & \underline{54.58} & \underline{62.78} & \textbf{52.90} & \textbf{52.90} & \underline{45.95} & \textbf{54.39} \\
\midrule
\multicolumn{13}{l}{\textit{\textbf{Native Proactive Streaming Models}}} \\
Em-Garde & \textbf{48.07} & \textbf{55.28} & 31.92 & 37.61 & 40.60 & 59.35 & 25.59 & 46.83 & 10.85 & 16.59 & 31.41 & 43.13 \\
JoyVL & 19.25 & 24.23 & 47.47 & \underline{54.39} & 19.66 & 50.62 & 43.15 & 52.19 & 24.92 & 24.92 & 30.89 & 41.27 \\
MMDuet2 & 0.36 & 0.44 & 26.00 & 26.00 & 0.00 & 0.00 & 17.00 & 17.00 & 1.46 & 1.46 & 8.96 & 8.98 \\
VideoChat3 & 14.47 & 28.01 & 20.00 & 21.79 & 25.21 & 26.07 & 21.32 & 22.94 & N/A & N/A & N/A & N/A \\
\bottomrule
\end{tabular}%
}
\parbox{\textwidth}{\vspace{2pt}\footnotesize Event averages the three Ego4D subsets equally; Overall averages the five tasks equally. Intervention B reuses U where no bidirectional semantic scores are available. N/A denotes VideoChat3 scores that cannot be recomputed without Intervention semantic scores. VideoChat3 uses its archived raw outputs, which are distinct from the existing main-table score artifacts.}
\endgroup
\end{table}

\FloatBarrier

\end{document}